%% file: main.tex
\documentclass[10pt,twocolumn,letterpaper]{article}
\usepackage{fofo-layout}
\input{preamble}
\usepackage{stfloats}
\title{Focus and Follow: Action-Guided Patch Allocation\\with Temporal Priors for Vision-Language-Action Models}
\author{%
  Yunkai Guo\textsuperscript{1,2}\quad Chenhui Shi\textsuperscript{1,3}\quad
  Genming Cui\textsuperscript{4}\quad Zonghao Guo\textsuperscript{1,2}\\
  Hao Wei\textsuperscript{1,3,*}\quad Haoran Jiang\textsuperscript{5}\quad
  Yihong Wu\textsuperscript{1,2,3,*}\\[5pt]
  {\normalsize\textsuperscript{1}State Key Laboratory of Multimodal Artificial Intelligence Systems, Institute of}\\
  {\normalsize Automation, Chinese Academy of Sciences, Beijing 100190, China}\\
  {\normalsize\textsuperscript{2}School of Artificial Intelligence, University of Chinese Academy of Sciences, Beijing 100049, China}\\
  {\normalsize\textsuperscript{3}YUKUN Intelligent World, Beijing, China}\\
  {\normalsize\textsuperscript{4}University of Science and Technology of China, Hefei, China}\quad
  {\normalsize\textsuperscript{5}Traffic Control Technology Co., Ltd., Beijing, China}\\[2pt]
  {\small\textsuperscript{*}\,Corresponding Authors}\\
  {\normalsize\ttfamily
    \href{mailto:guoyunkai2025@ia.ac.cn}{guoyunkai2025@ia.ac.cn}\quad
    \href{mailto:yhwu@nlpr.ia.ac.cn}{yhwu@nlpr.ia.ac.cn}\quad
    \href{mailto:weihao2019@ia.ac.cn}{weihao2019@ia.ac.cn}}}
\hypersetup{
  pdftitle={Focus and Follow: Action-Guided Patch Allocation with Temporal Priors for Vision-Language-Action Models},
  pdfauthor={Yunkai Guo, Chenhui Shi, Genming Cui, Zonghao Guo, Hao Wei, Haoran Jiang, Yihong Wu},
  urlcolor=fofogreen,
  pdflang={en}
}
\begin{document}
\raggedbottom
\maketitle
\suppressfloats[t]
\input{sections/motivation_figure}
\input{sections/comparison_figure}
\input{sections/abstract}
\input{sections/introduction}
\input{sections/architecture_figure}
\input{sections/related_work}
\input{sections/grid_generation_figure}
\input{sections/method}
\input{sections/experiments}
\input{sections/discussion}
\label{main:lastpage}
\clearpage
{\small\input{main.bbl}}
\clearpage
\appendix
\setcounter{figure}{0}
\setcounter{table}{0}
\renewcommand{\thefigure}{S\arabic{figure}}
\renewcommand{\thetable}{S\arabic{table}}
\renewcommand{\theHfigure}{supp.\arabic{figure}}
\renewcommand{\theHtable}{supp.\arabic{table}}
\numberwithin{equation}{section}
\renewcommand{\dbltopfraction}{.92}
\renewcommand{\textfraction}{.05}
\makeatletter
\setlength{\@fptop}{0pt}
\setlength{\@dblfptop}{0pt}
\makeatother
\twocolumn[{\pdfbookmark[0]{Supplementary Material}{supplement}%
  \begin{center}\Large\bfseries Supplementary Material\end{center}\vspace{6pt}}]
\input{appendix}

\end{document}

%% file: preamble.tex
\usepackage[T1]{fontenc}
\usepackage{array}
\usepackage{microtype,xcolor,xurl}
\usepackage{colortbl}
\definecolor{foforowgray}{gray}{0.93}
\definecolor{linkblue}{rgb}{0.21,0.49,0.74}
\definecolor{fofogreen}{RGB}{56,113,81}
\usepackage[pagebackref,breaklinks,colorlinks,allcolors=linkblue]{hyperref}
\DeclareMathOperator{\RowNorm}{RowNorm}
\graphicspath{{figures/}}

%% file: sections/motivation_figure.tex
\begin{figure}[t]
\centering
\includegraphics[width=\columnwidth,height=0.62923001\columnwidth,keepaspectratio]{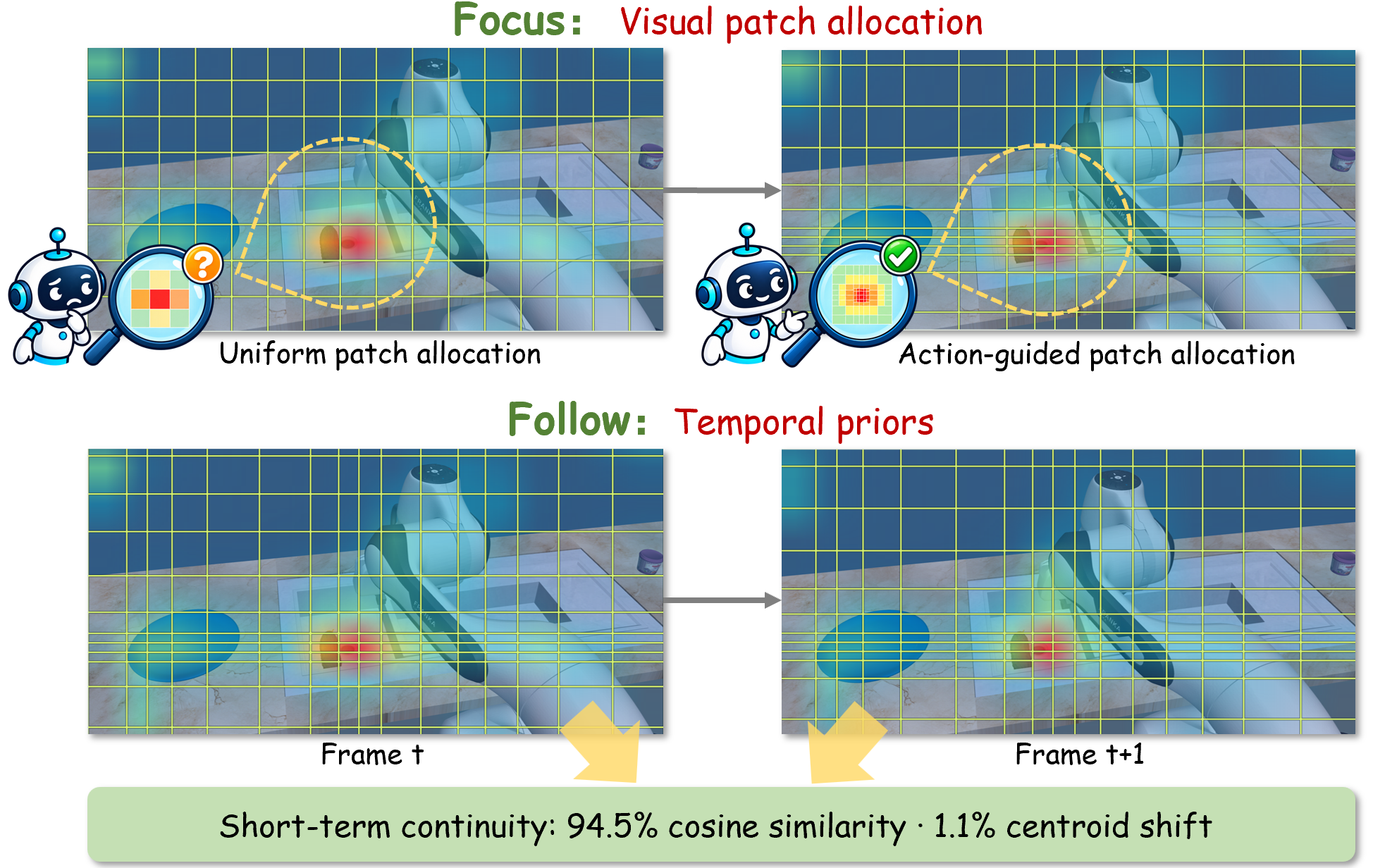}
\caption{\textbf{Action-guided patch allocation with temporal priors.}
\textbf{Top: Focus} uses action-guided image patchification to redistribute a fixed patch budget before encoding.
\textbf{Bottom: Follow} uses historical action-to-visual attention to guide the next frame's patch allocation.
The illustrated adjacent attention maps show 94.5\% cosine similarity and 1.1\% centroid shift, demonstrating short-term continuity.}
\label{fig:motivation}
\end{figure}

%% file: sections/comparison_figure.tex
\begin{figure*}[t]
\centering
\includegraphics[width=\textwidth]{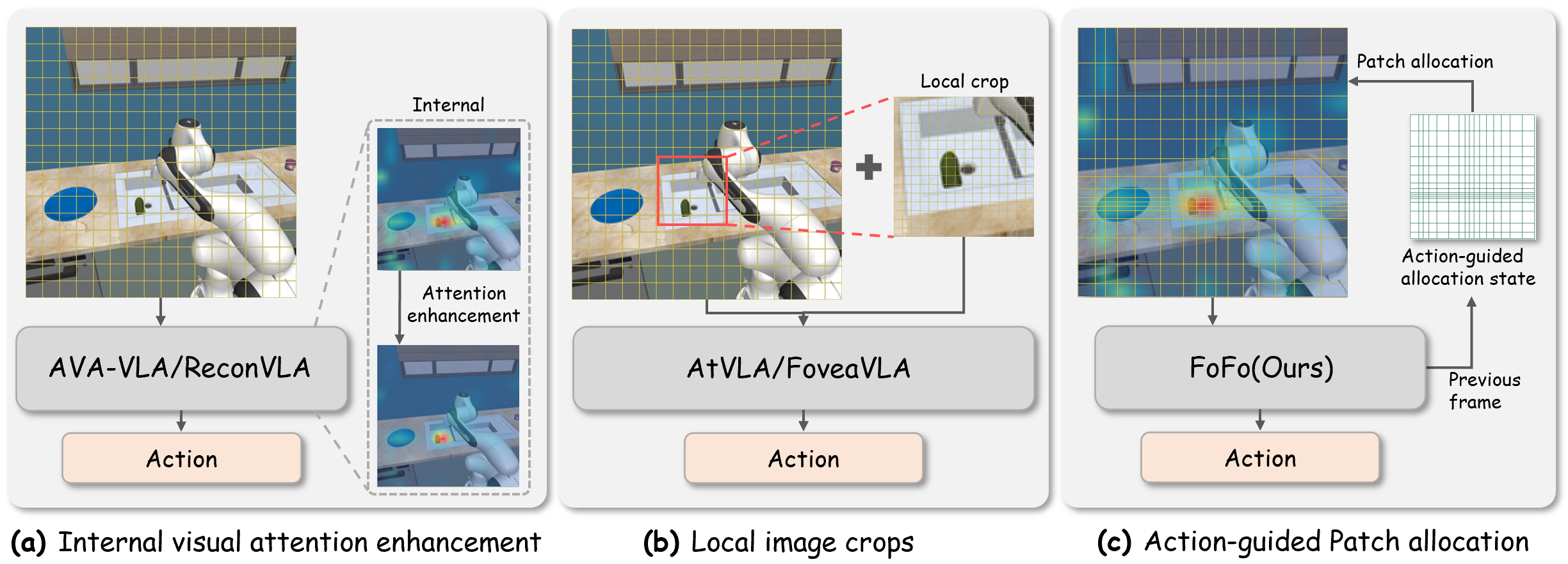}
\caption{\textbf{Comparison of task-aware visual processing strategies.}
\textbf{(a) Internal visual attention enhancement} uses explicit modulation or auxiliary supervision on encoded patch features, leaving the initial patch allocation unchanged.
\textbf{(b) Local image crops} supplement the global observation with local views, increasing the total patch count.
\textbf{(c) Action-guided patch allocation (FoFo)} uses an action-guided allocation state derived from the previous frame's action-to-visual attention to allocate current-frame patches before visual encoding, with the total patch count fixed.}
\label{fig:method_comparison}
\end{figure*}

%% file: sections/abstract.tex
\begin{abstract}
Recently, vision-language-action (VLA) models have demonstrated exceptional capability in completing specified tasks. However, most VLAs allocate the same patch density to task-relevant regions as to the background. Existing solutions attempt to implicitly strengthen attention of encoded patch features or explicitly increase patches of task-relevant regions through local image cropping, but the former leaves the initial patch allocation unchanged and the latter needs a substantial computational cost. In addition, we observe that the internal attention distribution of the VLA exhibits short-term continuity along manipulation trajectories. Based on these insights, we propose Focus and Follow (FoFo), an action-guided patch density allocation framework that uses historical attention as a temporal prior. Specifically, Focus uses an action-guided allocation state to assign more patches to task-relevant regions before visual encoding while preserving full-image coverage. Follow uses the historical frame\textquoteright{}s attention to determine the allocation state of the current frame and repeats this process at each step. We further introduce a training strategy, which reuses the allocation state of the current frame to provide temporal priors for two adjacent frames along the trajectory, avoiding repeated historical-attention computation. Extensive evaluations in simulation and real-world environments demonstrate that FoFo can significantly improve model success rates with only a few fine-tuning steps and 2.6 ms of additional inference delay, while achieving a significant $5.31\times$ training speedup over historical-attention recomputation.
\par\noindent Project page: \href{https://focus-and-follow.github.io/}{\normalfont focus-and-follow.github.io/}.
\end{abstract}

%% file: sections/introduction.tex
\section{Introduction}
Vision-language-action (VLA) models combine visual observations and language instructions to generate robot actions, having demonstrated strong capabilities across manipulation tasks~\cite{openvla,openvlaoft,pi05}. Pretrained vision-language representations and diverse robot data support adaptation through fine-tuning and generalization to unfamiliar environments~\cite{openvla,pi05}. As these capabilities expand, effective use of visual information becomes increasingly important for reliable action generation.

Most current VLAs assign the same initial patch density to task-relevant regions and the background (Fig.~\ref{fig:motivation}, top left), although critical regions may occupy little image area. Existing solutions implicitly strengthen attention to encoded patch features through modulation or auxiliary supervision~\cite{avavla,reconvla}, leaving the initial patch allocation unchanged (Fig.~\ref{fig:method_comparison}a). Others explicitly add task-relevant patches through local image cropping (Fig.~\ref{fig:method_comparison}b), increasing computation relative to the global view alone~\cite{atvla,foveavla,fisheyevla}. These limitations lead to our central question: \emph{How can we assign more patches to task-relevant regions before each frame is encoded, while preserving full-image coverage and keeping the total patch count unchanged?}

Inspired by attention- and saliency-guided sampling in other vision tasks~\cite{attwarp,fovea,salisa,learningtozoom}, we adopt \emph{nonuniform patchification in source images}: increasing task-relevant patch density and reducing background density under a fixed budget (Fig.~\ref{fig:motivation}, top right). The required \emph{allocation state} can be derived from internal attention. However, obtaining this attention before encoding poses another challenge: current-frame attention requires a model forward pass. We observe short-term continuity in internal attention distributions along manipulation trajectories: adjacent decisions attend to similar regions (Fig.~\ref{fig:motivation}, bottom). Based on this observation, we propose \textbf{Focus and Follow (FoFo)}, an action-guided patch density allocation framework that uses historical attention as a temporal prior to provide the allocation state in advance (Fig.~\ref{fig:method_comparison}c).

Specifically, \emph{Focus} uses an action-guided allocation state to assign more patches to task-relevant regions before visual encoding while preserving full-image coverage and the total patch count. It changes source-image patch density while retaining the existing visual encoder~\cite{siglip,dinov2}, and remains complementary to internal visual attention enhancement. \emph{Follow} uses the historical frame's action-to-visual attention to determine the allocation state of the current frame. Current action generation supplies attention for the next allocation, and this process repeats at each step without a separate attention-estimation pass.

To improve training efficiency, we introduce an \emph{allocation-state reuse} training strategy. It reuses the allocation state derived from the current frame's attention to provide temporal priors for two adjacent frames along the trajectory, avoiding repeated historical-attention computation. Randomly interleaving samples across trajectories preserves sampling randomness.

Extensive simulation experiments show that FoFo improves success rates across VLA models with different architectures. In particular, a few fine-tuning steps on $\pi_{0.5}$ yield an increase of 15.6 percentage points in success rate, with only 2.6\,ms of additional inference latency per decision. The supporting allocation-state reuse training strategy achieves a $5.31\times$ training speedup over historical-attention recomputation. Real-robot experiments further demonstrate FoFo\textquoteright{}s applicability in real-world settings.

Our main contributions are as follows:
\begin{itemize}
\setlength{\itemsep}{0pt}
\setlength{\parsep}{0pt}
\item We propose Focus and Follow (FoFo), an action-guided patch density allocation framework that uses historical attention as a temporal prior. Focus uses an action-guided allocation state to reallocate patches within a fixed budget before visual encoding while preserving full-image coverage, and Follow uses the allocation state produced at the preceding decision to avoid an additional attention computation.
\item We introduce an efficient training strategy which reuses the allocation state of the current frame to provide temporal priors for two adjacent frames along the trajectory, avoiding repeated historical-attention computation.
\item Extensive evaluations in simulation and real-world environments demonstrate that FoFo achieves a significant improvement in success rate at a very low cost, and it is adaptable to various models and real-world scenarios.
\end{itemize}

%% file: sections/architecture_figure.tex
\begin{figure*}[t]
\centering
\includegraphics[width=.85\textwidth,height=0.56822107\textwidth,keepaspectratio]{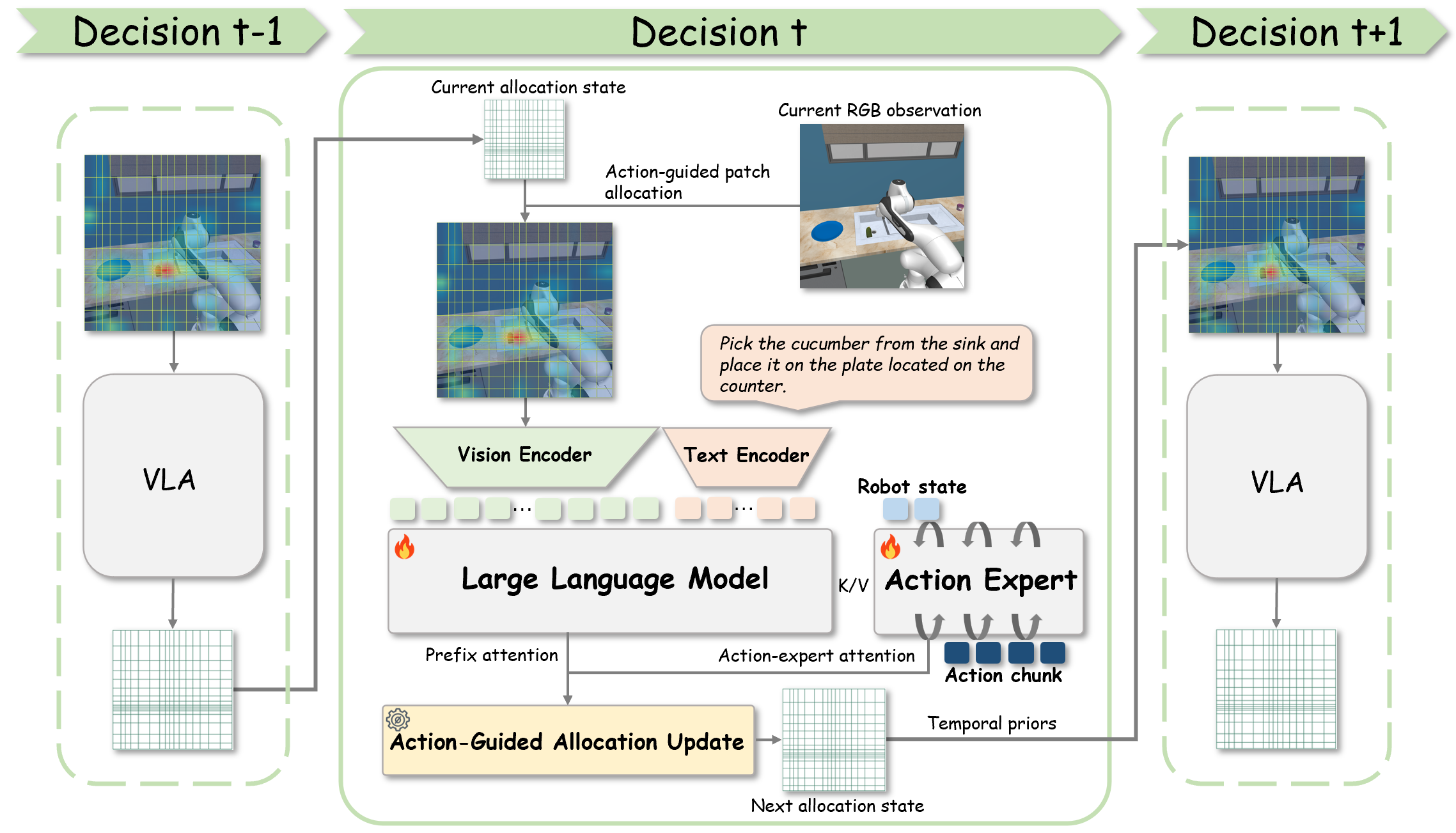}
\caption{\textbf{FoFo: action-guided patch allocation with temporal priors.} Focus uses the current allocation state to allocate patches before visual encoding without increasing their total count. Vision and text encoders supply the language model; its prefix K/V and robot state condition the action expert to produce an action chunk. \emph{Action-Guided Allocation Update} combines prefix and action-expert attention to compute the next allocation state. Follow carries this state to the next decision as a temporal prior. Flames mark trainable policy branches.}
\label{fig:recurrence}
\end{figure*}

%% file: sections/related_work.tex
\section{Related Work}
\paragraph{Vision-language-action Models.}
OpenVLA fuses DINOv2 and SigLIP patch features for language-model-based action prediction~\cite{openvla}. OpenVLA-OFT adds parallel decoding, action chunking, and continuous L1 regression, supporting separately encoded input views~\cite{openvlaoft}. $\pi_{0.5}$ combines a PaliGemma backbone with a flow-matching action expert and SigLIP visual encoding~\cite{pi05}. Their standard interfaces form visual sequences from regular patch grids. FoFo is compatible with these policy designs, redistributing source-image samples before encoding, within the original visual-token budget and interface.

\paragraph{Task-Aware Visual Processing in Robot Policies.}
Internal visual attention enhancement uses explicit modulation or auxiliary supervision to focus on important regions, without changing initial visual patch allocation (Fig.~\ref{fig:method_comparison}a). AVA-VLA combines recurrent state and language-conditioned visual features to modulate encoded visual-token attention~\cite{avavla}. ReconVLA guides attention through auxiliary diffusion supervision that reconstructs VAE latents of automatically localized target regions~\cite{reconvla}. Spatial Forcing aligns intermediate visual embeddings with pretrained 3D representations~\cite{spatialforcing}.

Local image crops supplement global observations with local views (Fig.~\ref{fig:method_comparison}b). AtVLA conditionally re-encodes attention-selected crops and appends local features to the cached prefix for action refinement~\cite{atvla}. FoveaVLA learns next-query interaction regions from region annotations and combines global and cropped external RGB-D inputs to replace wrist-camera observations~\cite{foveavla}. Fisheye-VLA combines a global fisheye view with calibrated local perspective crops around the end-effectors, matching the front-and-two-wrist baseline's token budget~\cite{fisheyevla}. ActiveVLA localizes task-critical regions in 3D and selects virtual viewpoints with 3D zoom~\cite{activevla}.

Look, Focus, Act learns foveated patch allocation from synchronized human gaze; Fov-Act jointly predicts future gaze and actions~\cite{lookfocusact}. FoFo uses policy-native action-to-visual attention for action-guided image patchification, preserving full-image coverage and the original token budget without additional local views, gaze supervision, or a separate current-attention estimation pass. Its spatial allocation complements internal visual attention enhancement, which leaves the input grid unchanged.

\paragraph{Temporal priors in robot policies.}
TraceVLA adds visual prompts of tracked point trajectories to express recent motion~\cite{tracevla}. HiF-VLA encodes historical motion vectors as tokens conditioning a joint action-and-motion expert~\cite{hifvla}. MemoryVLA retrieves and fuses perceptual details and semantic summaries from a consolidated memory bank~\cite{memoryvla}. VLA-Cache reuses historical key--value representations for stable, task-irrelevant visual tokens while recomputing changing or task-relevant ones~\cite{vlacache}. These approaches enrich policy context or reuse encoded representations. FoFo retains historical action-to-visual attention and sampling geometry as a temporal prior, sampling and encoding fresh RGB while current action generation updates the next allocation state. Training-state reuse likewise avoids repeated attention computation for historical frames.

%% file: sections/grid_generation_figure.tex
\begin{figure}[t]
\centering
\includegraphics[width=.80\columnwidth,height=1.53797980\columnwidth,keepaspectratio]{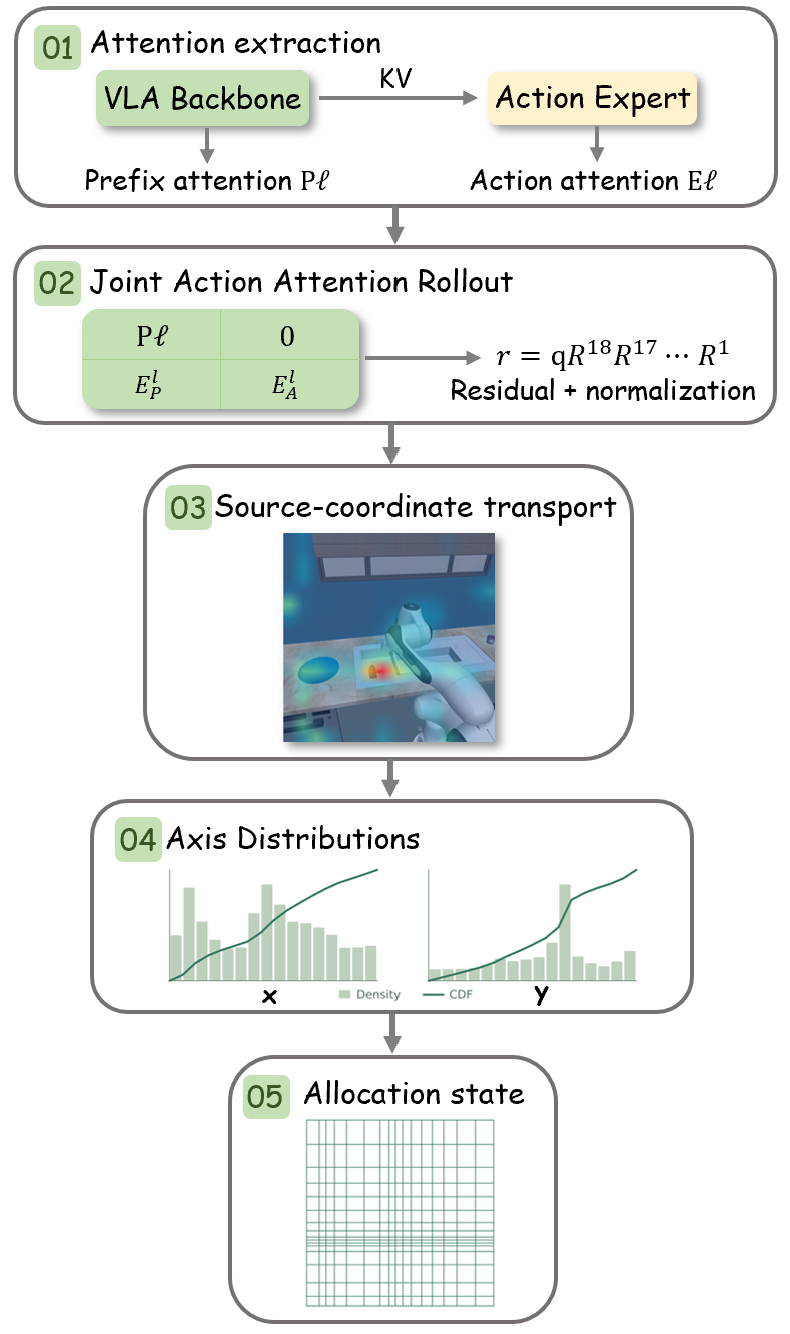}
\caption{\textbf{Action-guided allocation update.} (1) Extract prefix and action-expert attention from the current decode. (2) Add residual connections, normalize, and perform joint action attention rollout (18 layers in $\pi_{0.5}$). (3) Transport the resulting action-to-visual attention into source coordinates. (4) Form axis distributions, mix in a uniform component, and form their CDFs. (5) Invert the CDFs to construct the allocation state's nonuniform grid. During training, the same supervised flow forward supplies attention, conditioned on noised target actions and flow time, without an auxiliary decode.}
\label{fig:grid_generation}
\end{figure}

%% file: sections/method.tex
\section{Method}
FoFo combines two operations (Fig.~\ref{fig:recurrence}): Focus performs
\emph{action-guided image patchification}, and Follow provides its temporal
prior through \emph{temporal allocation reuse}.
Figure~\ref{fig:grid_generation} details the action-guided allocation update.

\subsection{Overview: action-guided patch allocation}
At decision $t$, the policy receives current RGB images $I_t$, robot state
$s_t$, and instruction $\ell$. Each trajectory maintains an allocation
state $H_t$ containing source-coordinate attention mass and the associated
sampling geometry. Starting from uniform sampling, FoFo follows
\begin{equation}
 \begin{aligned}
 \widetilde I_t &= \operatorname{Sample}(I_t,H_t),\\
 (A_t,M_t^{\mathrm{out}}) &=
 \operatorname{Decode}_{\theta}(\widetilde I_t,s_t,\ell),\\
 H_{t+1} &= \operatorname{Update}_{\alpha}(M_t^{\mathrm{out}},H_t).
 \end{aligned}
 \label{eq:recurrence}
\end{equation}
Here $A_t$ is the action chunk and $M_t^{\mathrm{out}}$ is the visual
attention mass extracted from the same decode. The \emph{current allocation
state} $H_t$ guides sampling before encoding; the \emph{next allocation
state} $H_{t+1}$ is used only at the next decision. The update transports
attention into source coordinates and constructs the next sampling map.

\subsection{Focus: action-guided image patchification}
\paragraph{Joint action attention rollout.}
We extract prefix attention and action-expert attention during the native
policy decode. Residual-aware attention rollout~\cite{attentionrollout}
traces action queries through both attention paths:
\begin{equation}
 \begin{aligned}
 R^l &= \RowNorm\!\left(
 \begin{bmatrix}P^l&0\\E_P^l&E_A^l\end{bmatrix}+\mathbf I\right),\\
 r &= qR^L R^{L-1}\cdots R^1.
 \end{aligned}
 \label{eq:action_rollout}
\end{equation}
Here $P^l$ is prefix attention, while $E_P^l$ and $E_A^l$ are action-query
attention over prefix and action keys. $\RowNorm$ normalizes each row after
adding the residual. The initial weights $q$ are uniform over the executed
action-query positions and zero elsewhere; $\pi_{0.5}$ uses the first five
positions and $L=18$ layers. Each camera's visual entries in $r$ reshape
into action-to-visual attention $M_t^{\mathrm{out}}$, directly connecting
patch allocation to action generation. Averaging details are in the supplement.

\paragraph{Source-coordinate transport.}
Attention is measured on the sampled canvas, whose patches can cover
different source-image regions at different decisions. We therefore
transport it into common source coordinates before reuse:
\begin{equation}
 M_t^{\mathrm{src}}=O_t^yM_t^{\mathrm{out}}(O_t^x)^\top.
 \label{eq:transport}
\end{equation}
Each entry of $O_t^x$ or $O_t^y$ is the fraction of a patch's consumed
source interval that overlaps a uniform source bin along that axis.
The transport preserves total attention mass under a uniform-within-patch
approximation, preventing output-patch indices from being mistaken for
fixed source locations.

\paragraph{Axis distributions and nonuniform grid.}
For each adaptive camera, we normalize $M_t^{\mathrm{src}}$ and sum it
along each axis to obtain attention marginals $a^x,a^y$. Mixing these with
uniform bin masses $u^x,u^y$ gives
\begin{equation}
 p^x=\alpha u^x+(1-\alpha)a^x,\qquad
 p^y=\alpha u^y+(1-\alpha)a^y.
 \label{eq:mixture}
\end{equation}
We use attention weight $\rho=1-\alpha=.65$ and uniform density floor
$\alpha=.35$; zero total attention uses uniform marginals.
Let $F^x,F^y$ be the cumulative distribution functions (CDFs) obtained
from piecewise-constant densities within source bins. Uniform canvas
coordinates $v=(v_x,v_y)\in[0,1]^2$ map to source coordinates through
\begin{equation}
 T(v)=\big((F^x)^{-1}(v_x),(F^y)^{-1}(v_y)\big).
 \label{eq:inversecdf}
\end{equation}
This \emph{nonuniform grid} places more samples in high-attention regions,
while the positive uniform component retains full-source coverage.
FoFo retains the original visual encoder and changes only the patch density
across source-image regions, with the total patch count fixed.
Sampling uses the full inverse CDF, bilinear lookup, and $2\times2$
subpixel averaging; implementation details are in the supplement.

\subsection{Follow: temporal allocation reuse}
Action-to-visual attention is available only after the policy processes an
observation. The Focus-only comparator therefore uses two policy decodes:
one estimates attention and the other generates actions
from the resampled image. Follow instead uses the preceding decision's
allocation as a temporal prior. It samples the current RGB image with
$H_t$, then obtains both the current action chunk and the attention needed
to construct $H_{t+1}$ in one decode (Eq.~\eqref{eq:recurrence}).
Thus each decision refreshes the allocation for the next observation
without an extra current-image attention-estimation pass.

Only allocation mass and sampling geometry persist across decisions;
current pixels are always encoded afresh, and prefix K/V serves only the
current decode. States are independent across trajectories, reset to
uniform sampling at episode start, and committed after successful action
output. In $\pi_{0.5}$, each ten-step decode predicts 50 actions and the
first five execute before the next decision. Short-term attention
continuity supports this reuse, but fast motion or a task transition can
make the prior stale. The uniform component preserves scene coverage
and limits sampling displacement when attention changes. The supplement
analyzes this sensitivity.

\subsection{Allocation-state reuse training strategy}
\label{sec:training}
The allocation-state reuse training strategy stores detached allocation states in a
bank indexed by trajectory and traversal direction. A randomly chosen
starting block uses uniform sampling; after successful optimization, its
next state is copied into independent forward and backward states.
Training randomly selects ready trajectories while preserving temporal
order within each traversal direction. Each visit samples its current
image using the corresponding stored state.

We obtain the next allocation from attention already computed in the
\emph{same native supervised flow forward} that produces the action loss.
This attention is conditioned on the noised target actions and flow time.
We detach it, apply joint action attention rollout and source-coordinate
transport, and store the result after a successful optimizer update.
Only the visited direction is updated; the other state is retained.
The new state serves a later visit, and gradients do not cross the bank.

Each visit therefore requires one supervised forward, without an auxiliary
policy decode or history-frame attention recomputation. Native action
targets, noise, flow time, normalization, padding, and loss are preserved;
no gaze, region, or grid supervision is added. Reverse traversal can use
states from later observations during training. Deployment remains causal
and updates its state from generated actions without target-action access.
The supplement further discusses the differences between this training
strategy and other training strategies.

%% file: sections/experiments.tex
\section{Experiments}
\subsection{Experimental Setup}
We evaluate $\pi_{0.5}$ and Xiaomi-Robotics-1 on RoboCasa365~\cite{robocasa365},
which tests 50 tasks: 18 Atomic-Seen, 16 Composite-Seen, and 16 Composite-Unseen.
Overall success is weighted by task count. Training uses the Human300
dataset, comprising 300 tasks with 100 human demonstrations per task.
We fine-tune $\pi_{0.5}$ for
5,000 steps on five NVIDIA RTX 6000 GPUs, with batch 64, LoRA rank 16,
learning rate $5\times10^{-6}$, and three $224\times224$ views.
Focus adapts the two external views after native preprocessing and retains
the native wrist view. Each view keeps $16\times16$ patch tokens, preserving
768 visual tokens in total. Each decision executes five actions.
The Xiaomi-Robotics-1 training configuration is provided in the supplement.

\subsection{Evaluation Results}
\input{sections/robocasa_table}
Our evaluation uses 50 complete rollouts per task under random seeds.
Table~\ref{tab:robocasa} shows that FoFo improves success across both VLA
architectures after a few fine-tuning steps. Overall success increases by
\textbf{15.6 percentage points} for $\pi_{0.5}$ and \textbf{2.5 points}
for Xiaomi-Robotics-1 relative to their published base models~\cite{robocasa_leaderboard}.
The $\pi_{0.5}$ gains span all three splits, with the largest improvement
on Atomic-Seen tasks (+20.7 points). Xiaomi-Robotics-1 also improves on
all splits and achieves the highest overall success among the methods
in the table. These results support action-guided patch allocation across
different VLA architectures while retaining their native visual interfaces.

\subsection{Ablation Studies}
\input{sections/component_ablation_table}
We study the components on RoboCasa's 18 Atomic-Seen tasks
(Table~\ref{tab:components}). All variants are initialized from pretrained
$\pi_{0.5}$ and trained for 800 steps on five NVIDIA RTX 6000 GPUs,
with all other conditions held identical. Native $\pi_{0.5}$ uses uniform sampling;
\emph{extra local crop} estimates attention and adds a local view;
\emph{Focus} reallocates a fixed patch budget using current-frame attention
from an extra decode; \emph{Focus+Follow} instead reuses historical attention;
\emph{full FoFo} further adds the allocation-state reuse training strategy.

\paragraph{Patch allocation.}
Focus improves success by 21.5 percentage points over native $\pi_{0.5}$
and by 7.5 points over extra local cropping. In the selected-frame references,
Focus assigns $2.59\times$ as many patches to key regions as uniform sampling,
consistent with the benefit of denser task-relevant visual input.
Local cropping adds 256 patches, increasing the total visual-patch budget by
33.3\%; Focus keeps the total fixed. ``Patches in task-relevant regions''
refers to the mean number of patches within task-relevant regions per frame.
Counting and sampling details are in the supplement.

\paragraph{Inference efficiency.}
Follow removes the extra attention-estimation decode, reducing inference
time by 51.6\% ($2.07\times$ speedup) relative to Focus alone, without
compromising the success rate. Full FoFo shares this inference path and
improves success by 21.7 points over native
$\pi_{0.5}$ with only about 2.6\,ms of additional latency per decision.

\paragraph{Training efficiency.}
While Follow can accelerate inference, it incurs additional time overhead
during training because the attention from the previous frame must be
computed. The allocation-state reuse strategy stores and reuses the
attention from each frame---utilizing a cleverly designed data
organization---thereby achieving a $5.31\times$ speedup in optimizer
updates compared to recomputing the history window, with an overhead of
only 9.1\,ms (7.7\%) relative to native training.
Thus the complete method improves task success while keeping inference
and training costs close to those of the base policy.

\subsection{Attention Weight Analysis}
\label{sec:attention_experiment}
\input{sections/attention_grid_figure}
\input{sections/attention_figure}
The attention weight $\rho=1-\alpha$ in Eq.~\eqref{eq:mixture} controls
the balance between attention-guided and uniform sampling. Larger $\rho$
concentrates more patches in attended regions (Fig.~\ref{fig:grid_sweep}).
We vary only this weight on the 18 Atomic-Seen tasks.
Relative to uniform sampling, success improves by 21.6 percentage points
near $\rho=.65$ (Fig.~\ref{fig:attention_sweep}), supporting that allocating
a higher density of patches to task-relevant regions can improve the
success rate. Increasing the weight to 1.00 instead reduces
success by 17.3 points from this peak: excessive concentration sacrifices
surrounding context needed for manipulation. We therefore use $\rho=.65$,
retaining a 35\% uniform component. Configuration details are in the supplement.

\subsection{Real-World Experiments}
\label{sec:real_robot}
\input{sections/real_robot_mechanism}
\input{sections/real_robot_success}
Our platform uses an AgileX PiPER six-degree-of-freedom arm~\cite{piper},
three Intel RealSense cameras: two external cameras and one wrist-mounted camera.
We collect 100 demonstration trajectories per task for part assembly,
banana placement into a basket, drawer opening, and tabletop tidying,
where multiple objects are sorted into different containers.
For each task, we evaluate each policy for 100 trials across different
object placements.

Table~\ref{tab:real_robot} shows success gains of 29, 21, 15, and 12
percentage points over $\pi_{0.5}$ across these four tasks, averaging
19.25 points. The largest gain occurs in assembly,
which requires precise perception of a small part and its target.
Figure~\ref{fig:real_robot_mechanism} illustrates denser patch allocation
in key interaction regions. The mean patch counts in task-relevant regions in Table~\ref{tab:real_robot}
are $3.16\times$, $2.44\times$, $2.72\times$, and $2.68\times$
those of the baseline, respectively, without increasing the total patch budget. Assembly has the
largest local increase, consistent with its demand for fine visual detail.
Together, these observations support FoFo's applicability to real-world
manipulation through better allocation of existing visual resources.

%% file: sections/robocasa_table.tex
\begin{table*}[t]
\caption{\textbf{FoFo improves RoboCasa (November 1, 2026)~\cite{robocasa_leaderboard} manipulation success across models.} FoFo achieves an improvement in success rates over both the classic method $\pi_{0.5}$ and the state-of-the-art (SOTA) method Xiaomi-Robotics-1, while requiring only a minimal number of additional fine-tuning steps---a testament to its practical value.}
\label{tab:robocasa}
\centering
\fontsize{7.5}{9}\selectfont
\setlength{\tabcolsep}{5pt}
\renewcommand{\arraystretch}{1.03}
\begin{tabular}{@{}llrrrr@{}}
\toprule
Policy & Family & \shortstack{Atomic\\Seen (18)} & \shortstack{Composite\\Seen (16)} & \shortstack{Composite\\Unseen (16)} & \shortstack{Overall\\(50)}\\
\midrule
Paimon-0 (Primotion)~\cite{paimon0} & VLA & 79.3 & 55.8 & \textbf{36.7} & 58.1\\
Xiaomi-Robotics-1 (Xiaomi Robotics)~\cite{xiaomi} & VLA & 80.2 & 57.1 & 32.1 & 57.4\\
Phasor-m7 (Phasor)~\cite{phasorm7} & VLA & 78.7 & 57.4 & 31.5 & 56.8\\
Meituan-Robotics-0 (Meituan Robotics)~\cite{meituan} & VLA & 77.3 & 57.8 & 27.1 & 55.0\\
LoopWAM (Guo et al.)~\cite{loopwam} & WAM & 78.0 & 48.3 & 15.4 & 48.5\\
ProWAM (Zhang et al.)~\cite{prowam} & WAM & 76.0 & 46.5 & 18.2 & 48.1\\
PatchWAM (Wang et al.)~\cite{patchwam} & WAM & 78.7 & 46.4 & 13.8 & 47.6\\
ABot-M0.5 (Chen et al.)~\cite{abotm05} & WAM & 75.6 & 37.7 & 3.3 & 40.3\\
PRTS (Zhang et al.)~\cite{prts} & VLA & 66.3 & 30.3 & 18.8 & 39.6\\
RLDX-1 (Kim et al.)~\cite{rldx1} & VLA & 67.6 & 27.9 & 8.5 & 36.0\\
WorldDreamer (World Agents)~\cite{worlddreamer} & WAM & 66.3 & 26.7 & 9.0 & 35.3\\
GR00T N1.5 (NVIDIA)~\cite{groot15} & VLA & 50.7 & 14.8 & 2.7 & 23.9\\
LY-GWM-RoboCasa (Ma et al.)~\cite{lygwm} & VLA + WM & 48.3 & 16.3 & 3.1 & 23.6\\
GR00T N1.6 (NVIDIA)~\cite{groot16} & VLA & 51.1 & 9.4 & 1.7 & 21.9\\
GigaWorld-Policy 0.1 (Ye et al.)~\cite{gigaworld} & WAM & 44.4 & 11.8 & 2.9 & 20.7\\
$\pi_{0.5}$ (Black et al.)~\cite{pi05} & VLA & 39.6 & 7.1 & 1.2 & 16.9\\
$\pi_0$ (Black et al.)~\cite{pi0} & VLA & 34.6 & 6.1 & 1.1 & 14.8\\
Diffusion Policy (Chi et al.)~\cite{diffusionpolicy} & Visuomotor & 15.7 & 0.2 & 1.3 & 6.1\\
\midrule
\rowcolor{foforowgray}
$\pi_{0.5}$ + \textbf{FoFo} & VLA & 60.3\,(+20.7) & 26.5\,(+19.4) & 7.3\,(+6.1) & 32.5\,(+15.6)\\
\rowcolor{foforowgray}
Xiaomi-Robotics-1 + \textbf{FoFo} & VLA & \textbf{84.1}\,(+3.9) & \textbf{59.5}\,(+2.4) & 33.1\,(+1.0) & \textbf{59.9}\,(+2.5)\\
\bottomrule
\end{tabular}
\end{table*}

%% file: sections/component_ablation_table.tex
\begin{table*}[t]
\caption{\textbf{Patch allocation improves success; temporal reuse reduces cost.} Compared to the ``extra local crop'' strategy, the ``Focus'' strategy provides more patches in task-relevant regions without increasing the total visual patches. This results in a higher success rate. Furthermore, the ``Follow'' strategy reduces inference time to a level comparable to that of $\pi_{0.5}$, significantly enhancing inference efficiency. The associated training strategy also maintains a training speed similar to that of $\pi_{0.5}$. In short, ``FoFo'' achieves a substantial improvement in success rate with virtually no increase in training or inference time.}
\label{tab:components}
\centering
\fontsize{8.5}{10}\selectfont
\setlength{\tabcolsep}{2.8pt}
\renewcommand{\arraystretch}{1.12}
\resizebox{\textwidth}{!}{%
\begin{tabular}{@{}lccccccrrrr@{}}
\toprule
Policy & Focus & Follow & \shortstack{State reuse\\training strategy} &
\shortstack{Patches in task-\\relevant regions\\(mean)} &
\shortstack{Total visual\\patches} &
\shortstack{Decodes/\\decision} & \shortstack{Success $\uparrow$\\(\%)} &
\shortstack{Inference $\downarrow$\\(ms)} & \shortstack{Train $\downarrow$\\(s/update)} &
\shortstack{Speedup $\uparrow$\\($\times$)}\\
\midrule
Native $\pi_{0.5}$ & No & No & No & 25.00 & 768 & 1 & 41.3 & 69.24 & 0.119 & 5.72\\
$\pi_{0.5}$ + extra local crop & No & No & No & 51.8 & 1024 & 2 & 55.3 & 259.42 & 0.355 & 1.92\\
$\pi_{0.5}$ + Focus & Yes & No & No & \textbf{64.67} & 768 & 2 & 62.8 & 148.59 & 0.305 & 2.24\\
$\pi_{0.5}$ + Focus + Follow & Yes & Yes & No & \textbf{64.67} & 768 & 1 & 62.9 & 71.86 & 0.681 & 1.00\\
\midrule
\rowcolor{foforowgray}
$\pi_{0.5}$ + FoFo (full) & Yes & Yes & Yes & \textbf{64.67} & 768 & 1 & \textbf{63.0} & 71.86 & 0.128 & 5.31\\
\bottomrule
\end{tabular}}
\end{table*}

%% file: sections/attention_grid_figure.tex
\begin{figure*}[t]
\centering
\includegraphics[width=.92\textwidth]{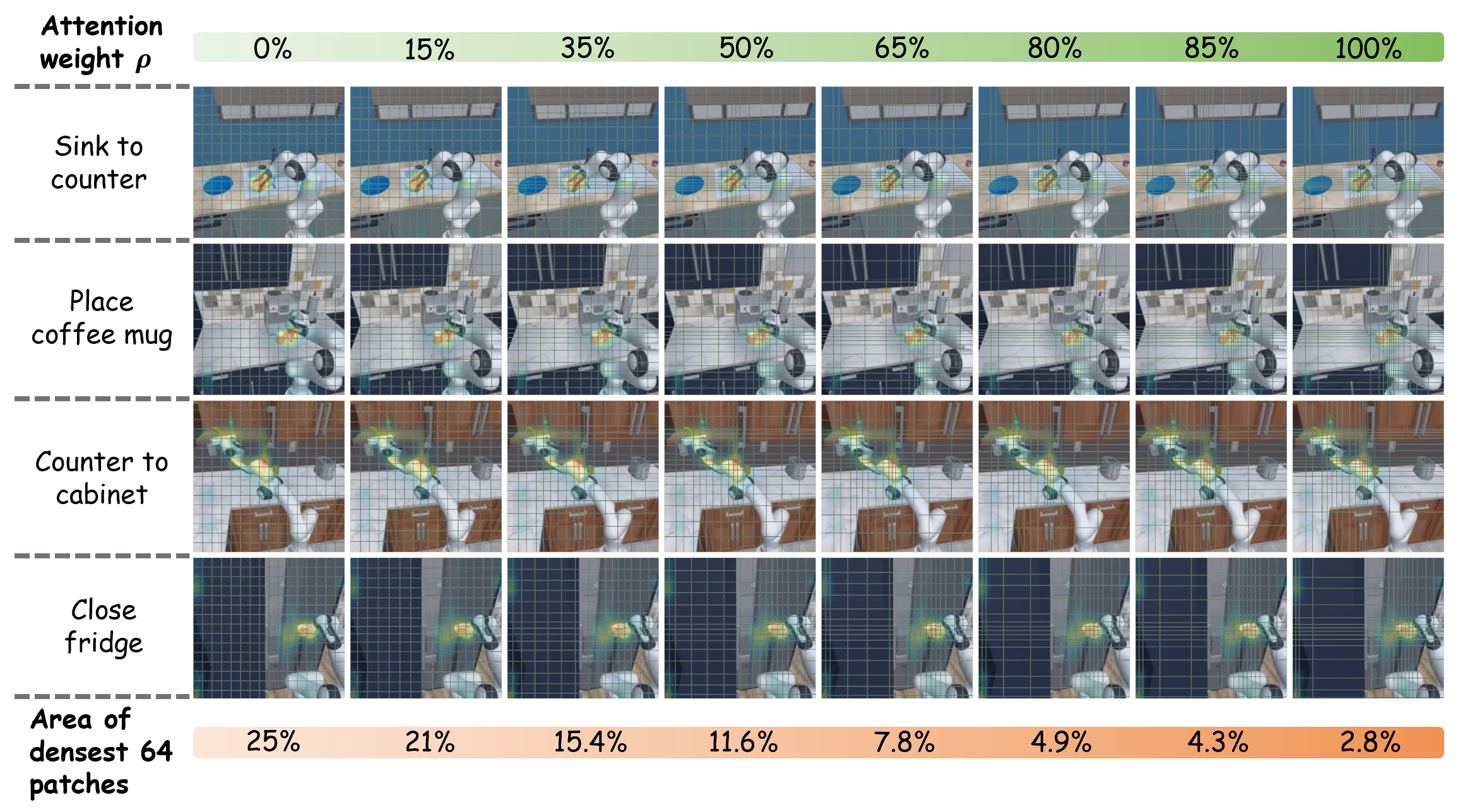}
\caption{\textbf{Attention weight controls patch concentration.} Increasing $\rho$ concentrates a fixed patch budget into smaller task-relevant regions, as shown by the shrinking area covered by the densest 64 patches. Stronger concentration allocates fewer patches to surrounding context.}
\label{fig:grid_sweep}
\end{figure*}

%% file: sections/attention_figure.tex
\begin{figure}[t]
\centering
\includegraphics[width=\linewidth]{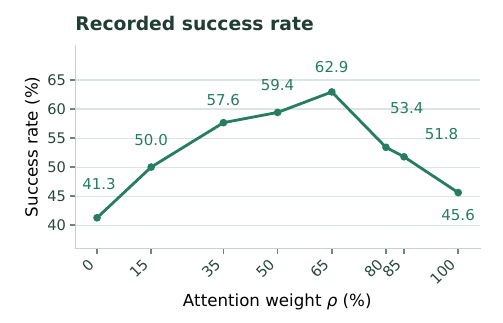}
\caption{\textbf{Moderate attention concentration improves success.} Success rises as $\rho$ increases toward .65, but declines when excessive concentration reduces sampling of surrounding context. The axes show attention weight $\rho$ and success rate; configuration details are in the supplement.}
\label{fig:attention_sweep}
\end{figure}

%% file: sections/real_robot_mechanism.tex
\begin{figure*}[t]
\centering
\includegraphics[width=\textwidth]{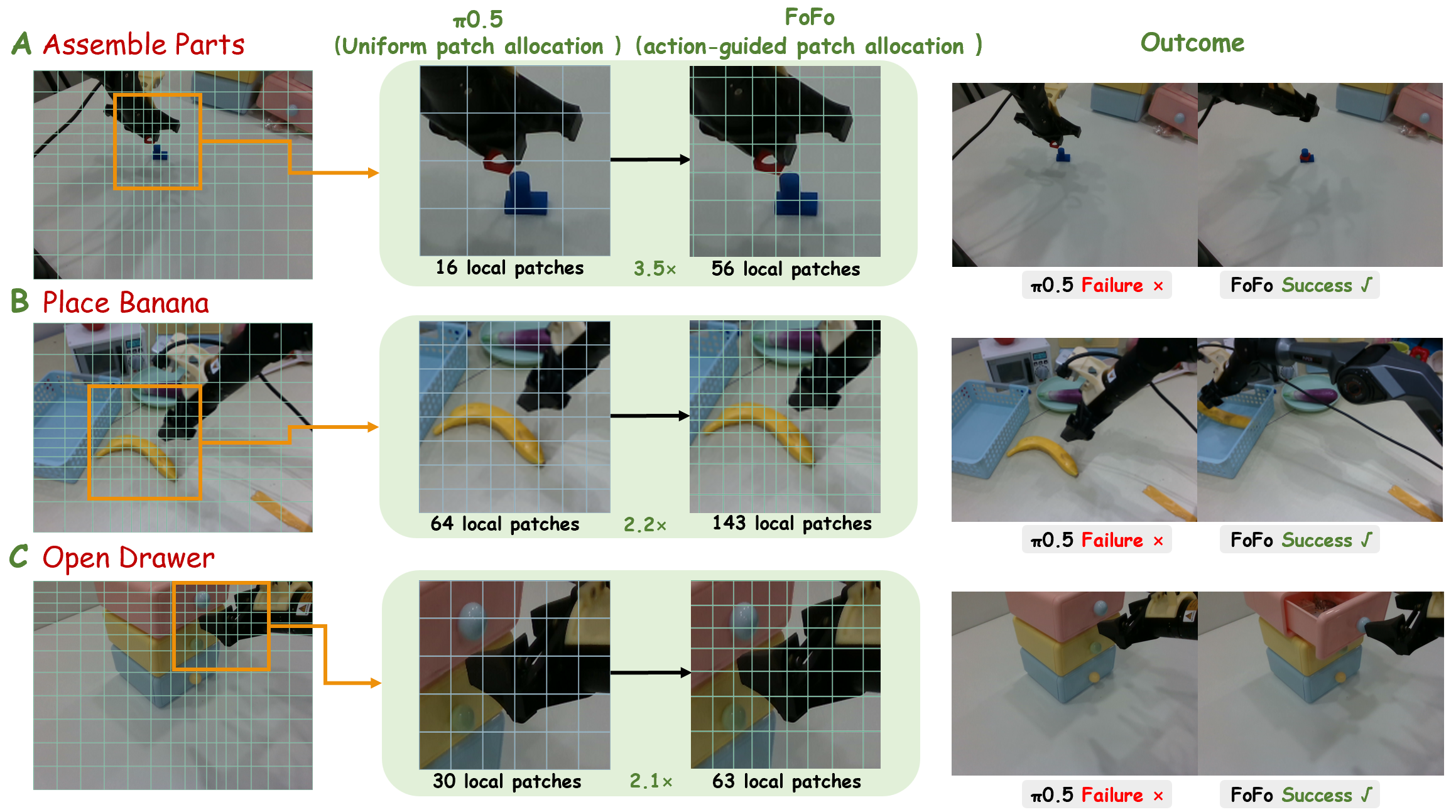}
\caption{\textbf{Action-guided allocation concentrates patches on real-world interaction regions.} A--C show Assemble Parts, Place Banana, and Open Drawer, with the same marked ROI under uniform and action-guided allocation. Denser local patches illustrate how FoFo allocates more visual resources to task-relevant regions. Table~\ref{tab:real_robot} reports mean patch counts in task-relevant regions and success gains.}
\label{fig:real_robot_mechanism}
\end{figure*}

%% file: sections/real_robot_success.tex
\begin{table}[t]
\centering
\caption{\textbf{FoFo improves real-robot success with denser local allocation.} The mean gain over $\pi_{0.5}$ is 19.25 percentage points, with the largest gain in assembly. SR: success rate (\%); P: mean patches in task-relevant regions.}
\label{tab:real_robot}
\small
\setlength{\tabcolsep}{3.5pt}
\renewcommand{\arraystretch}{1.18}
\begin{tabular}{@{}l*{8}{r}@{}}
\toprule
Method & \multicolumn{2}{c}{\shortstack{Assemble\\Parts}} & \multicolumn{2}{c}{\shortstack{Place\\Banana}} & \multicolumn{2}{c}{\shortstack{Open\\Drawer}} & \multicolumn{2}{c}{\shortstack{Tidy\\Table}} \\
\cmidrule(lr){2-3}\cmidrule(lr){4-5}\cmidrule(lr){6-7}\cmidrule(l){8-9}
 & SR & P & SR & P & SR & P & SR & P \\
\midrule
$\pi_{0.5}$ & 48 & 25 & 73 & 25 & 68 & 25 & 55 & 25 \\
FoFo & \textbf{77} & \textbf{79} & \textbf{94} & \textbf{61} & \textbf{83} & \textbf{68} & \textbf{67} & \textbf{67} \\
\bottomrule
\end{tabular}
\end{table}

%% file: sections/discussion.tex
\section{Conclusion and future work}
We identify a visual patch allocation problem in VLAs: uniform sampling
assigns the same initial patch density to regions with different task
relevance. FoFo addresses this bottleneck before visual encoding by using
historical action-to-visual attention as a temporal prior. Focus redistributes a fixed patch
budget, Follow avoids a separate attention-estimation pass, and the
allocation-state reuse training strategy avoids repeated historical-attention computation.
Together, these components improve the use of existing visual resources,
increase task success, and reduce inference and training costs relative
to their recomputation baselines. Evaluations with $\pi_{0.5}$ and
Xiaomi-Robotics-1, together with real-robot experiments, support FoFo's
applicability across models and in physical manipulation.

Future work will explore extending action-guided patch allocation to
rapidly changing viewpoints, including wrist cameras. Effectively handling
significant inter-frame variations would facilitate a more comprehensive
utilization of historical attention information as a temporal prior.
Another direction is learning an adaptive attention weight to replace the
current fixed setting and adjust patch density to changing visual demands
across manipulation stages.

%% file: appendix.tex
This supplement provides additional algorithm details
(Sec.~\ref{app:impl}), mechanism analyses (Sec.~\ref{app:mechanisms}),
core settings and additional results (Sec.~\ref{app:protocols}), visual
analysis (Sec.~\ref{app:visual}), and real-world data collection
(Sec.~\ref{app:data_collection}). Main-paper figures, tables, equations,
and literature citations retain their original numbering;
additional sources are linked where cited.

\input{sections/implementation}
\input{sections/mechanism_studies}
\input{sections/protocols}

\input{sections/visualizations}
\input{sections/data_collection}
\input{sections/compact_evidence_page}
\input{sections/integrated_visual_figures}

%% file: sections/implementation.tex
\section{Additional Algorithm Details}
\label{app:impl}

\subsection{Attention averaging}
For $\pi_{0.5}$, let $P^{l,h}$ be prefix attention at layer $l$ and head $h$, and
$E^{l,h,s}$ the action-expert attention at flow step $s$. We average heads
within each branch and average the expert over the one-based steps
$\mathcal S=\{3,6,9\}$ of the ten-step action decode:
\begin{equation}
 P^l=\frac1{n_P}\sum_hP^{l,h},\qquad
 E^l=\frac1{|\mathcal S|n_E}\sum_{s\in\mathcal S}\sum_h E^{l,h,s}.
 \label{eq:app_attention_mean}
\end{equation}
Here $n_P$ and $n_E$ are the numbers of attention heads in the prefix
and action-expert branches, respectively. These means enter the
residual-normalized joint rollout in the main
paper~\cite{attentionrollout}. The resulting visual mass is normalized
per camera before constructing its allocation. During training, the
supervised forward supplies a head mean at its sampled flow time
(Sec.~\ref{app:training}).

\subsection{Inverse-CDF sampling and antialiasing}
For source-bin boundaries $b_j,b_{j+1}$, probability $p_j$, and cumulative
mass $c_j=\sum_{i<j}p_i$, the complete inverse CDF is
\begin{equation}
 Q(u)=b_j+\frac{u-c_j}{p_j}(b_{j+1}-b_j),
 \quad c_j\leq u<c_j+p_j.
 \label{eq:app_quantile}
\end{equation}
The map preserves the source endpoints. The uniform component $\alpha=1-\rho=.35$
gives each bin positive mass. We evaluate the full inverse CDF at each subpixel
sample and average $2\times2$ bilinear samples per output pixel. The
$224\times224$ output preserves the native $16\times16=256$ patches per view.

\subsection{Sensitivity of the temporal prior}
\label{app:geometry}
On a normalized source axis, let $A$ be the attention CDF,
$F_A(x)=\alpha x+\rho A(x)$ the mixed CDF, and $Q_A=F_A^{-1}$. For continuous piecewise-linear attention CDFs $A,B$ and $\alpha>0$,
\begin{equation}
 \|Q_A-Q_B\|_\infty\leq
 \frac{\rho}{\alpha}\|A-B\|_\infty.
 \label{eq:inversebound}
\end{equation}
To see this, the uniform component gives
$F_A(y)-F_A(x)\geq\alpha(y-x)$ for $y\geq x$.
At a common quantile $u$, take $x=Q_A(u)$ and $y=Q_B(u)$.
Then $\alpha|x-y|\leq|F_A(y)-F_B(y)|
\leq\rho\|A-B\|_\infty$, proving the bound.

For the default mixture, $\rho/\alpha=13/7$. The same bound applies to
patch boundaries and pixel samples. If consecutive attention CDFs
change by at most $\delta$, a lag-$d$ prior changes sampling coordinates
by at most $d\rho\delta/\alpha$. Using the most recent prior therefore
limits accumulated drift, while the uniform component stabilizes the
map and preserves scene coverage. Section~\ref{app:temporal} examines
this temporal structure in recorded trajectories.

\paragraph{Relation to prior sampling methods.}
AttWarp~\cite{attwarp} uses attention between the current image and question
to warp that same image. FoFo exploits the short-term continuity of attention
during manipulation. The allocation state derived from action attention at the
previous decision serves as the temporal prior for patchifying the current observation.
\mbox{SALISA}~\cite{salisa} builds saliency maps from preceding detection boxes and
scores and pretrains its sampler on annotation-derived maps. FoFo obtains the
attention for the next allocation directly from the policy during action
generation, without additional spatial annotations or an external attention expert.

%% file: sections/mechanism_studies.tex
\section{Additional Mechanism Analyses}
\label{app:mechanisms}

\subsection{Training with allocation-state reuse}
\label{app:training}
History reconstruction is the natural training formulation of Focus
and Follow: preceding observations are decoded to reconstruct the
allocation for each supervised observation. Our bidirectional
allocation-state reuse strategy builds on this formulation by retaining
detached states for each trajectory and traversal direction, avoiding
repeated history decoding. Single-forward reuse further extracts the
next attention from the supervised forward computing the action loss,
eliminating the current auxiliary decode. This attention is conditioned
on noised targets at the sampled flow time; collecting it preserves the
supervised objective and its gradients.

\subsection{Temporal structure of action attention}
\label{app:temporal}
We compare uniform allocation and standard one-decision reuse across 144
complete trajectories on the 18 Atomic-Seen tasks using the same policy weights.
We normalize attention in source coordinates per camera. For lags one through ten,
we use common observations and average first within and then across tasks.

\begin{figure}[t]
\centering
\includegraphics[width=\linewidth]{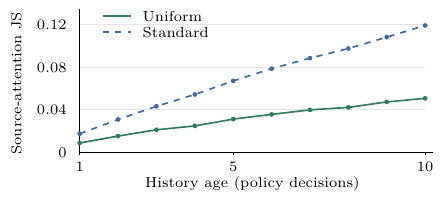}
\caption{\textbf{Recent attention provides a structured temporal prior.}
Task-balanced source-attention JS divergence on 72 complete trajectories
per mode. A lag unit is one policy decision, corresponding to five
executed actions. Both curves use common anchors across lags.}
\label{fig:temporal}
\end{figure}

In Figure~\ref{fig:temporal}, the increase in JS divergence from lag 1 to lag
10 is $.04203$ (95\% interval $[.03910,.04504]$) under uniform allocation
and $.10185$ ($[.09597,.10786]$) under reuse. We resample whole trajectories
within each task to estimate the intervals. In both settings, the most recent
decision better reflects the current spatial distribution.

\subsection{Spatial allocation at fixed policy weights}
\label{app:fixed_weights}
On the 18 RoboCasa Atomic-Seen tasks, we hold each policy's weights and
visual budget fixed and switch patchification at inference. FoFo fine-tuning
uses the main ablation configuration for 800 updates. Both policies follow
the main evaluation protocol with multiple seeds. Action-guided patchification
improves both the original Human300 $\pi_{0.5}$ and the FoFo-adapted policy,
with a larger gain after adaptation (Table~\ref{tab:same_weights_allocation}).
\input{sections/allocation_intervention_table}

\subsection{Source of the allocation signal}
\label{app:allocation_source}
All four variants train for 800 updates on the 18 RoboCasa Atomic-Seen
tasks using the same training settings and evaluation protocol as the
main-paper ablation. Only the patch allocation
signal varies (Table~\ref{tab:allocation_source}). Center focus assigns more
patches to the image center. Language-to-vision attention measures how strongly
instruction tokens attend to visual patches. All-prefix attention aggregates all valid prefix queries for each patch.
Action-to-vision attention traces action queries back to patches across layers.
\input{sections/attention_sources_table}

Pooling all prefix queries can direct patches toward regions unrelated to the
current interaction. With a fixed instruction, language attention can lose
track of the target as manipulation progresses. Action-guided attention ties
allocation to the motion being generated and achieves the highest success
at the same patch budget, exceeding the strongest alternative by 5.5
percentage points.

%% file: sections/allocation_intervention_table.tex
\begin{table}[t]
\centering\small
\caption{\textbf{Spatial allocation at fixed policy weights.} Each row compares success rates for two inference grids on the same policy.}
\label{tab:same_weights_allocation}
\setlength{\tabcolsep}{3pt}
\begin{tabular}{@{}lrr@{}}
\toprule
Model & \shortstack{Uniform\\patchification} & \shortstack{Action-guided\\patchification}\\
\midrule
Fine-tuned & 41.9\% & \textbf{61.2\%}\\
Not fine-tuned & 41.3\% & \textbf{52.9\%}\\
\bottomrule
\end{tabular}
\end{table}

%% file: sections/attention_sources_table.tex
\begin{table}[t]
\centering\small
\caption{\textbf{Allocation signals on RoboCasa Atomic-Seen.}}
\label{tab:allocation_source}
\begin{tabular}{@{}lr@{}}
\toprule
Allocation signal & Success rate\\
\midrule
Center focus & 49.5\%\\
Language-to-vision attention & 56.7\%\\
All prefix tokens $\to$ vision & 57.1\%\\
Action-to-vision attention & \textbf{62.6\%}\\
\bottomrule
\end{tabular}
\end{table}

%% file: sections/protocols.tex
\section{Core Settings and Additional Results}
\label{app:protocols}

\subsection{RoboCasa adaptation settings}
For RoboCasa, we adapt $\pi_{0.5}$ with rank-16 LoRA and train the native
projection heads. Each trajectory starts with uniform patchification and
uses the preceding decision's allocation thereafter.

\paragraph{Xiaomi-Robotics-1 adaptation.}
We adapt the public Xiaomi-Robotics-1 RoboCasa365 model~\cite{xiaomi}
on the full Human300 training split. We use rank-16 LoRA on the language
and action transformers and train the native projection heads.
Training uses a global batch of 40 and 10,000 updates. We retain the
native four-frame, three-camera input at $256\times256$.
Following the \href{https://github.com/XiaomiRobotics/Xiaomi-Robotics-1/blob/0dd7aef8dc87296246aae812a1f59ccb708e5546/eval_robocasa365/entry.py}{official RoboCasa protocol},
the policy predicts 16 actions and executes all 16 before replanning.
FoFo adapts the two most recent external-view frames; older frames and
wrist views retain their native representation. The mixture is
$\rho=.65$, with a causal allocation state along each demonstration.

\subsection{LIBERO implementation and results}
We follow the official openpi $\pi_{0.5}$ LIBERO recipe for
full-parameter fine-tuning, initializing from $\pi_{0.5}$-base and
training on \texttt{physical-intelligence/libero} for 30,000 updates
with a global batch size of 256 and a peak learning rate of
$5\times10^{-5}$.\footnote{Official openpi
\href{https://github.com/Physical-Intelligence/openpi/blob/215abfb217dbac7d5f1273282331b9b1866c0479/src/openpi/training/config.py}{training configuration}
and \href{https://github.com/Physical-Intelligence/openpi/blob/215abfb217dbac7d5f1273282331b9b1866c0479/examples/libero/main.py}{LIBERO evaluator}; $\pi_{0.5}$ results from the \href{https://github.com/Physical-Intelligence/openpi/blob/215abfb217dbac7d5f1273282331b9b1866c0479/examples/libero/README.md}{official benchmark report}.}
FoFo adapts the external agentview image and preserves the native wrist
view, with 256 visual tokens per view. The policy predicts ten 7D
actions and executes five before replanning; each trajectory maintains
its own allocation state. Each suite contains ten tasks. We evaluate
50 rollouts per task for each seed and report suite success rates
averaged over multiple evaluation seeds.

\begin{table}[t]
\centering\small
\caption{\textbf{LIBERO success rates (\%).} Published baseline results
retain their reported input and training protocols. FoFo attains the
highest success in each suite among the compared methods.}
\label{tab:libero}
\setlength{\tabcolsep}{2.3pt}
\begin{tabular}{@{}lrrrrr@{}}
\toprule
Method & Obj. & Spa. & Goal & Long & Avg.\\
\midrule
Diffusion Policy~\cite{openvla} & 92.5 & 78.3 & 68.3 & 50.5 & 72.4\\
Octo~\cite{openvla} & 85.7 & 78.9 & 84.6 & 51.1 & 75.1\\
OpenVLA~\cite{openvla} & 88.4 & 84.7 & 79.2 & 53.7 & 76.5\\
SpatialVLA\footnotemark & 89.9 & 88.2 & 78.6 & 55.5 & 78.1\\
$\pi_0$~\cite{openvlaoft} & 98.8 & 96.8 & 95.8 & 85.2 & 94.2\\
OpenVLA-OFT~\cite{openvlaoft} & 98.4 & 97.6 & 97.9 & 94.5 & 97.1\\
$\pi_{0.5}$~\cite{pi05} & 98.2 & 98.8 & 98.0 & 92.4 & 96.9\\
\midrule
FoFo & \textbf{99.6} & \textbf{99.1} & \textbf{99.1} & \textbf{97.1} & \textbf{98.7}\\
\bottomrule
\end{tabular}
\end{table}
\footnotetext{Qu et al., \emph{SpatialVLA}, RSS 2025.
\href{https://www.roboticsproceedings.org/rss21/p011.html}{Published paper}.}

Across the four LIBERO suites,\footnote{Liu et al., \emph{LIBERO}, NeurIPS 2023.
\href{https://libero-project.github.io/}{Benchmark and paper}.} FoFo averages 98.7\% success, 1.6 points above OpenVLA-OFT
(Table~\ref{tab:libero}). On Long, its 97.1\% success gives a 2.6-point gain.

\subsection{Xiaomi-Robotics-1 under a matched training budget}
Table~\ref{tab:xiaomi_matched} compares two adaptations of Xiaomi-Robotics-1
with 10,000 updates on Human300 and matched initialization, other training
settings, and evaluation. FoFo improves Atomic-Seen, Composite-Seen, and
Composite-Unseen success by 4.0, 2.1, and 1.2 percentage points, respectively.
\input{sections/xiaomi_matched_table}

%% file: sections/xiaomi_matched_table.tex
\begin{table}[t]
\centering\small
\caption{\textbf{Xiaomi-Robotics-1 success rates (\%) after the same number of training updates.}}
\label{tab:xiaomi_matched}
\setlength{\tabcolsep}{3pt}
\begin{tabular}{@{}lrrr@{}}
\toprule
Training method & \shortstack{Atomic\\Seen} & \shortstack{Composite\\Seen} & \shortstack{Composite\\Unseen}\\
\midrule
Native fine-tuning & 80.1 & 57.4 & 31.9\\
FoFo fine-tuning & \textbf{84.1} & \textbf{59.5} & \textbf{33.1}\\
\bottomrule
\end{tabular}
\end{table}

%% file: sections/visualizations.tex
\section{Visual Analysis of Patch Allocation}
\label{app:visual}
\subsection{ROI selection and local patch counting}
\label{app:roi_counting}
The region of interest (ROI) is a task-relevant source-image region used
to count local visual patches. In the supplementary automatic-ROI analysis, we place a
$48\times48$-pixel window with the largest task-relevance weight on the
$224\times224$ source canvas. All methods use the same selected camera
and ROI.
In Figure~\ref{fig:allocation}, the ROI is the illustrated local region.

Each patch is represented by its footprint in source-image coordinates.
We count it once whenever this footprint overlaps the ROI with positive
area, including partially covered patches. Contact along an edge alone
does not count. Uniform patches use the regular source grid; adaptive
patches use the actual sampling map consumed at that decision.
Per-frame counts are averaged over the evaluated observations.

\paragraph{Additional local crop.}
The crop baseline retains the global views and adds a $16\times16$
patch grid from one attention-selected crop. We map each crop patch back
to its source image using the actual crop rectangle and add its ROI
intersections to those of the global view. Only patches from the ROI's
source camera contribute. Global-view and crop-view tokens are counted
separately even when their source footprints overlap. Thus, enlarging
the crop changes its patch footprints but leaves the counting rule unchanged.

In the three examples in Figure~\ref{fig:allocation}, local patch counts
increase to $2.2\times$, $1.9\times$, and $3.6\times$ those under uniform
allocation, respectively, at the same global budget. This illustrates how spatial allocation concentrates
visual resources around the manipulation region.

\subsection{Simulation and real-world visualizations}
\label{app:measured_geometry}
Figures~\ref{fig:measured_geometry} and~\ref{fig:real_progress}(b) show how
the temporal prior maps to the grid and policy input on LIBERO and real-world
observations, respectively. Smaller source cells allocate more patches to the
manipulation region, while larger cells preserve the surrounding scene.

%% file: sections/data_collection.tex
\section{Real-World Data Collection}
\label{app:data_collection}
We collect demonstrations by teleoperating an AgileX PiPER arm~\cite{piper}
with a PIKA interface.\footnote{AgileX official
\href{https://github.com/agilexrobotics/pika_ros}{PIKA ROS documentation}
and \href{https://github.com/agilexrobotics/PikaAnyArm}{PikaAnyArm teleoperation software}.}
Two PIKA Station positioning base stations track one PIKA Sense
handheld unit (Figure~\ref{fig:collection_setup}); handheld motion and
gripper input guide the robot. Two external cameras record the
workspace and a wrist camera records local interactions. Camera streams
and robot commands form the demonstrations.

\subsection{Language and action interfaces}
\label{app:realrobot}
Table~\ref{tab:real_instructions} gives the original task instructions.
Each recorded action contains six joint targets and the gripper
opening, on a 50\,ms dataset time grid. The policy predicts 50 actions
with ten denoising steps and executes the first five before replanning.
Figure~\ref{fig:real_progress}(a) presents five chronological frames per
task, preserving the full camera view. The tidying sequence covers
both placements in instruction order. The numbers below the images
indicate decision steps; each step corresponds to five actions.

%% file: sections/compact_evidence_page.tex
\begin{figure*}[p]
\centering
\input{sections/roi_figure}
\par\vspace{7pt}
\input{sections/collection_setup_figure}
\par\vspace{7pt}
\input{sections/real_instructions_table}
\end{figure*}

%% file: sections/roi_figure.tex
\begin{minipage}{\textwidth}
\centering
\includegraphics[width=.74\textwidth]{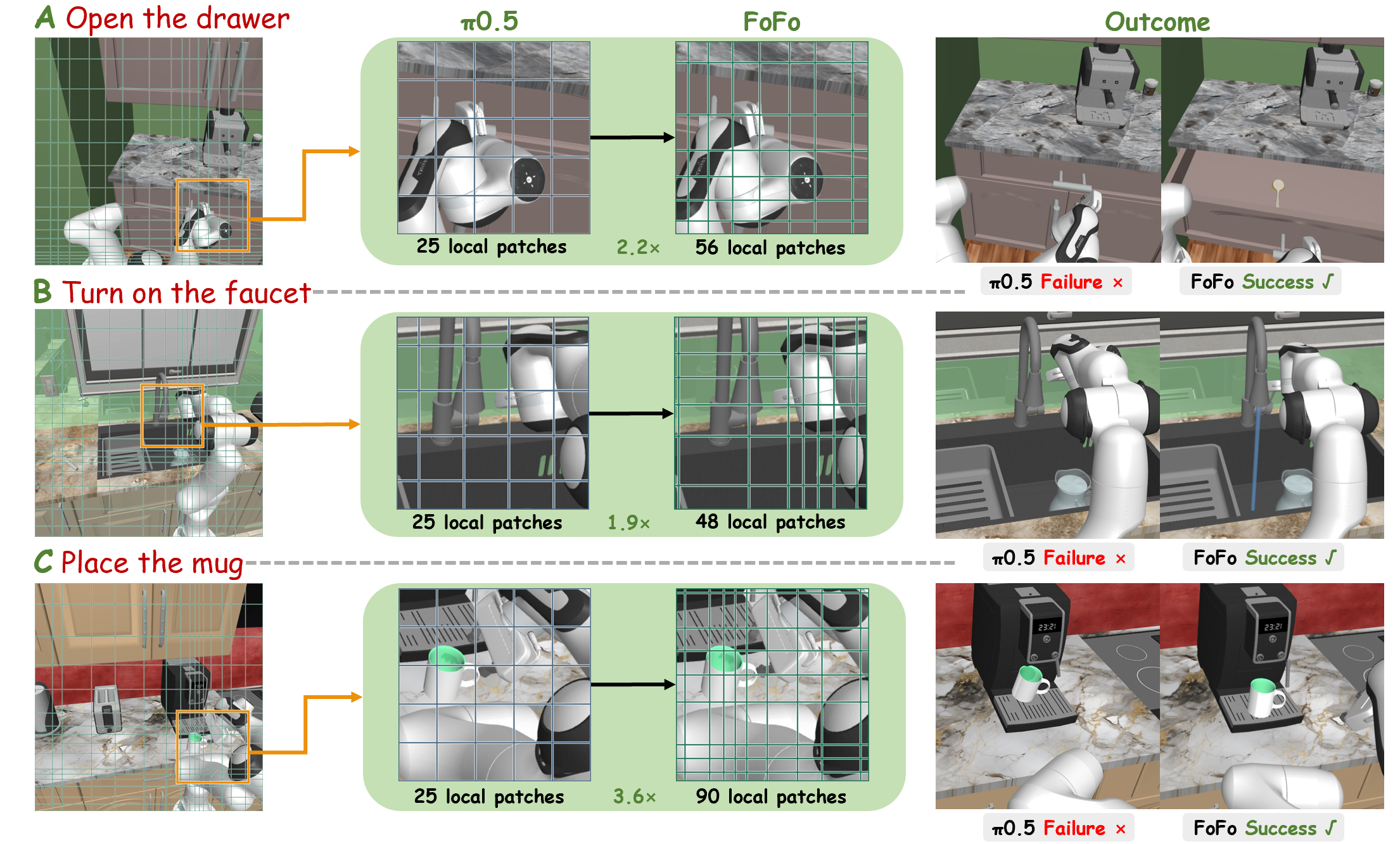}
\captionof{figure}{\textbf{Denser local allocation at a fixed global budget.}
FoFo examples compare adaptive and uniform grids
on the same RGB image. The drawer, faucet, and mug ROIs intersect the source footprints of
56, 48, and 90 adaptive patches, respectively, versus 25 uniform patches each.
Each view retains 256 patches.}
\label{fig:allocation}
\end{minipage}

%% file: sections/collection_setup_figure.tex
\begin{minipage}{\textwidth}
\centering
\includegraphics[width=.55\linewidth]{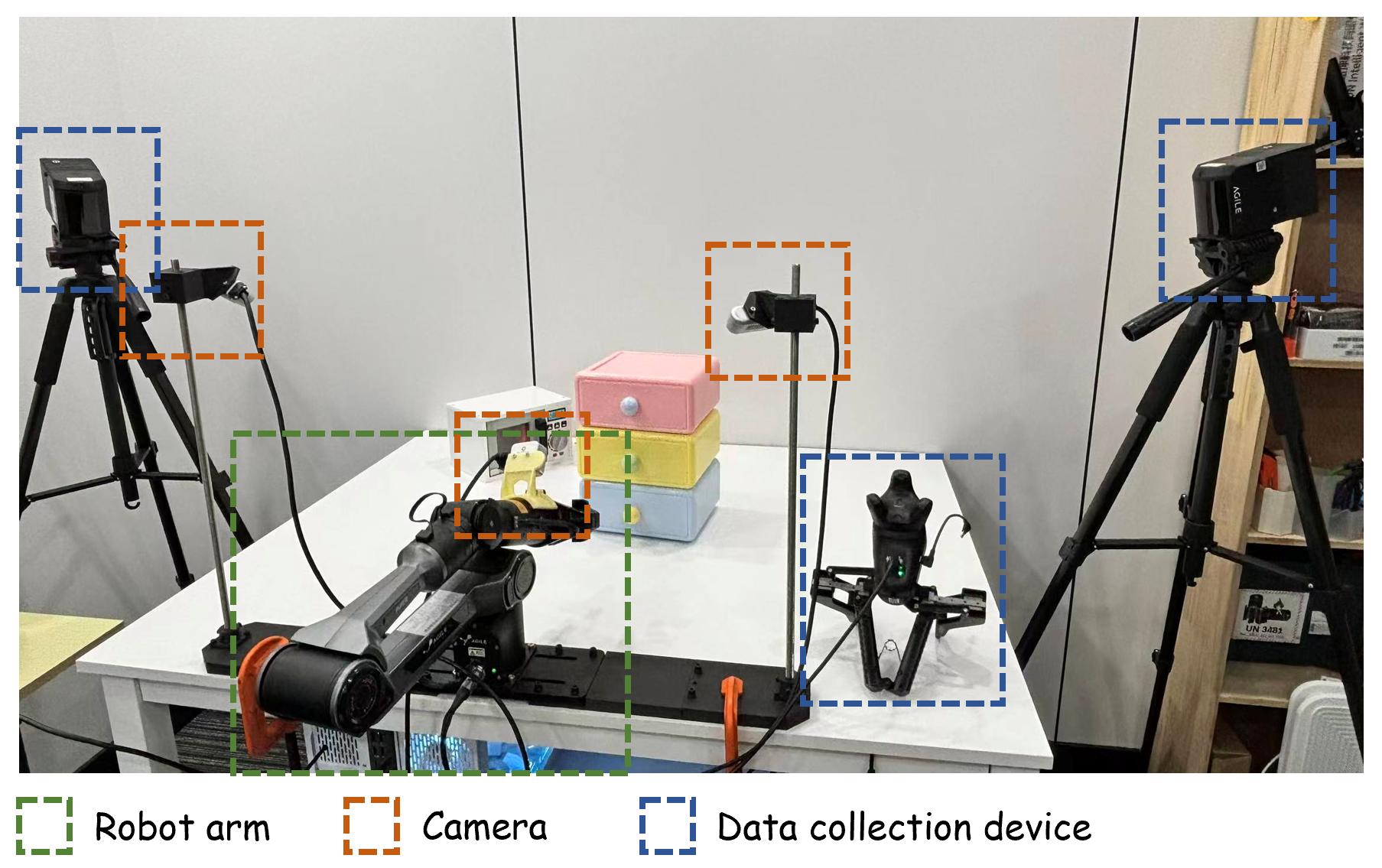}
\captionof{figure}{\textbf{Real-world demonstration collection setup.}
Green: PiPER arm. Orange: two external cameras and a wrist camera.
Blue: two PIKA Station positioning base stations and one PIKA Sense
handheld unit.}
\label{fig:collection_setup}
\end{minipage}

%% file: sections/real_instructions_table.tex
\begin{minipage}{\textwidth}
\centering\small
\captionof{table}{\textbf{Task instructions for the real-world demonstrations.}}
\label{tab:real_instructions}
\setlength{\tabcolsep}{5pt}
\renewcommand{\arraystretch}{1.15}
\begin{tabular}{@{}lp{.75\textwidth}@{}}
\toprule
Task & Recorded instruction\\
\midrule
Assemble Parts & Place the red part onto the blue peg.\\
Place Banana & Put the banana in the basket\\
Open Drawer & Open the pink middle drawer.\\
Tidy Table & First place the purple eggplant in the pink plate, then place the green vegetable in the blue plate.\\
\bottomrule
\end{tabular}
\end{minipage}

%% file: sections/integrated_visual_figures.tex
\begin{figure*}[p]
\centering
\includegraphics[width=\textwidth]{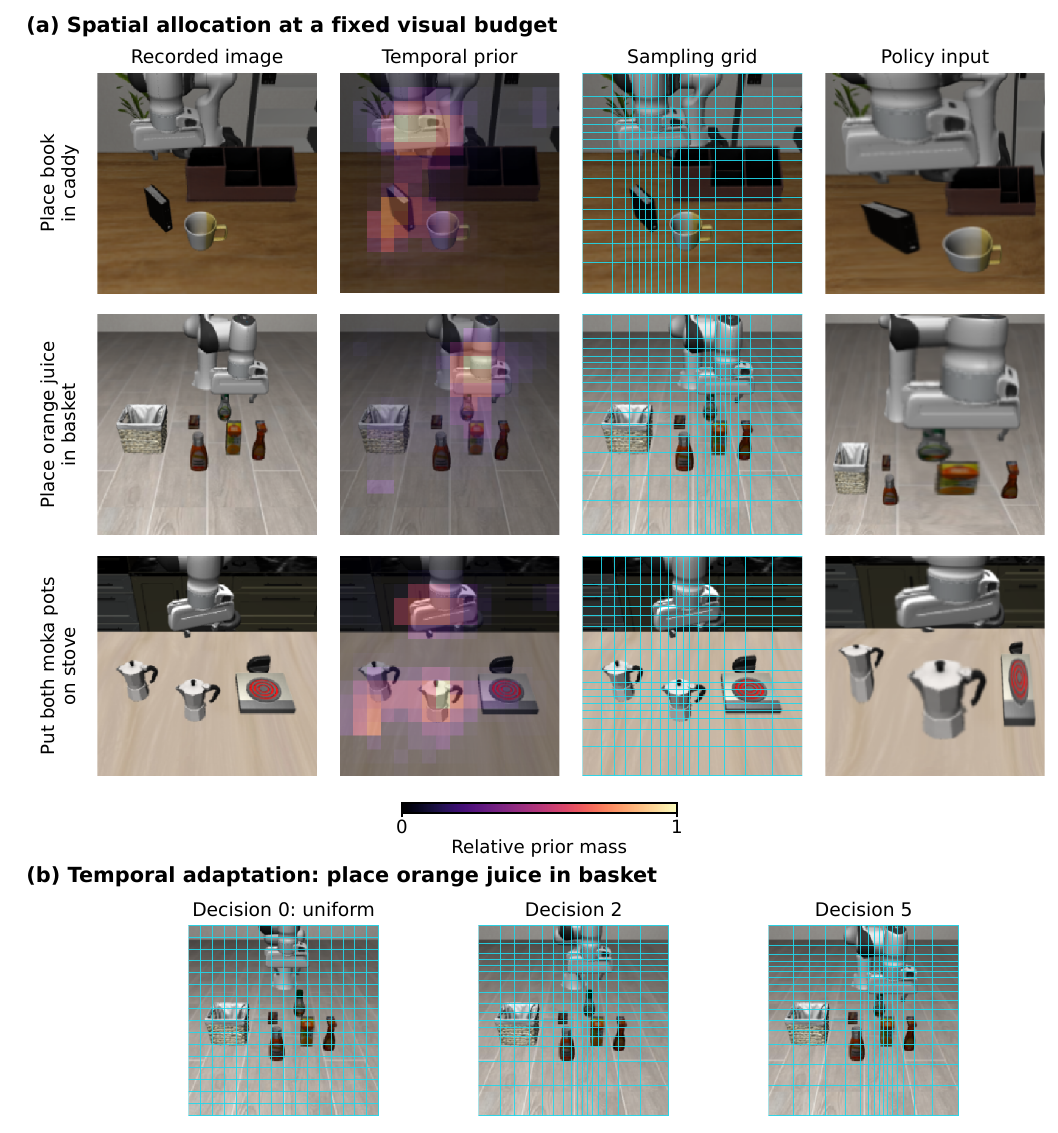}
\caption{\textbf{Spatial focus and temporal adaptation on LIBERO.}
(a) Examples from three tasks connect the recorded image, temporal prior,
source-image patch allocation, and policy input. (b) The orange-juice
sequence at decisions 0, 2, and 5 starts uniformly and then uses
preceding policy attention. Each view retains 256 patches.}
\label{fig:measured_geometry}
\end{figure*}

\begin{figure*}[p]
\centering
\includegraphics[width=\textwidth]{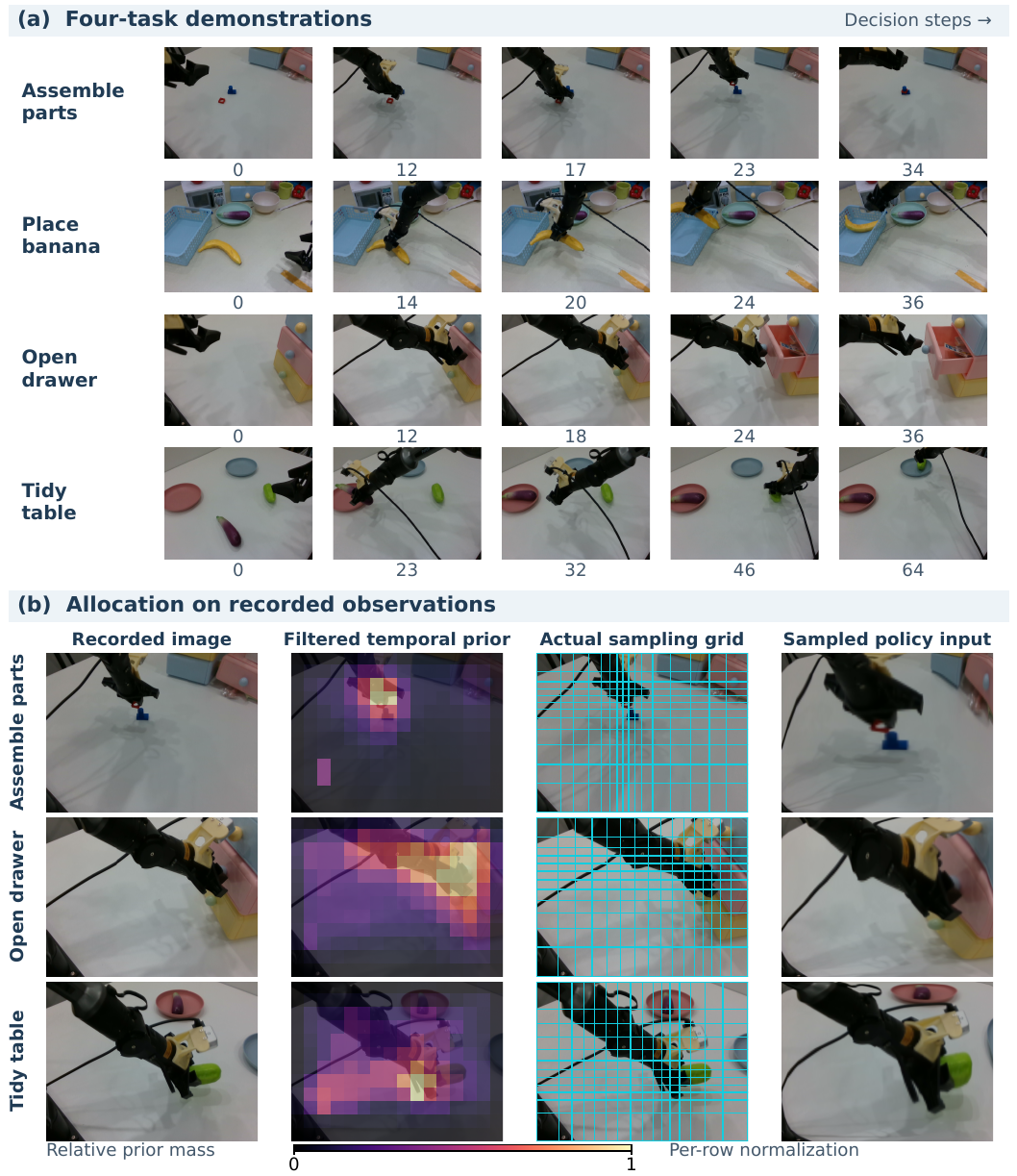}
\caption{\textbf{Four-task real-world demonstrations and visual allocation.}
(a) Five recorded frames per task, ordered left to right with complete
camera views. The numbers below the images indicate decision steps,
with five actions per step; tidying covers both placements. (b) Recorded images, temporal
priors, source-image patch allocations, and policy inputs. Each view
retains 256 patches.}
\label{fig:real_progress}
\end{figure*}

%% file: main.bbl
\begin{thebibliography}{42}
\providecommand{\natexlab}[1]{#1}
\providecommand{\url}[1]{\texttt{#1}}
\expandafter\ifx\csname urlstyle\endcsname\relax
  \providecommand{\doi}[1]{doi: #1}\else
  \providecommand{\doi}{doi: \begingroup \urlstyle{rm}\Url}\fi

\bibitem[Abnar and Zuidema(2020)]{attentionrollout}
Samira Abnar and Willem Zuidema.
\newblock Quantifying attention flow in transformers.
\newblock In \emph{Proceedings of the 58th Annual Meeting of the Association
  for Computational Linguistics}, pages 4190--4197, 2020.

\bibitem[{AgileX Robotics}()]{piper}
{AgileX Robotics}.
\newblock {PiPER}: 6-{DoF} lightweight robotic arm for manipulation and
  physical {AI}.
\newblock Official product page.
\newblock Accessed November 12, 2026.

\bibitem[Black et~al.(2025{\natexlab{a}})Black, Brown, Darpinian, Dhabalia,
  Driess, Esmail, Equi, Finn, Fusai, Galliker, Ghosh, Groom, Hausman, Ichter,
  Jakubczak, Jones, Ke, LeBlanc, Levine, Li-Bell, Mothukuri, Nair, Pertsch,
  Ren, Shi, Smith, Springenberg, Stachowicz, Tanner, Vuong, Walke, Walling,
  Wang, Yu, and Zhilinsky]{pi05}
Kevin Black, Noah Brown, James Darpinian, Karan Dhabalia, Danny Driess, Adnan
  Esmail, Michael~Robert Equi, Chelsea Finn, Niccolo Fusai, Manuel~Y. Galliker,
  Dibya Ghosh, Lachy Groom, Karol Hausman, Brian Ichter, Szymon Jakubczak, Tim
  Jones, Liyiming Ke, Devin LeBlanc, Sergey Levine, Adrian Li-Bell, Mohith
  Mothukuri, Suraj Nair, Karl Pertsch, Allen~Z. Ren, Lucy~Xiaoyang Shi, Laura
  Smith, Jost~Tobias Springenberg, Kyle Stachowicz, James Tanner, Quan Vuong,
  Homer Walke, Anna Walling, Haohuan Wang, Lili Yu, and Ury Zhilinsky.
\newblock {$\pi_{0.5}$}: A vision-language-action model with open-world
  generalization.
\newblock In \emph{Proceedings of the 9th Conference on Robot Learning}, pages
  17--40, 2025{\natexlab{a}}.

\bibitem[Black et~al.(2025{\natexlab{b}})Black, Brown, Driess, Esmail, Equi,
  Finn, Fusai, Groom, Hausman, Ichter, Jakubczak, Jones, Ke, Levine, Li-Bell,
  Mothukuri, Nair, Pertsch, Shi, Smith, Tanner, Vuong, Walling, Wang, and
  Zhilinsky]{pi0}
Kevin Black, Noah Brown, Danny Driess, Adnan Esmail, Michael~Robert Equi,
  Chelsea Finn, Niccolo Fusai, Lachy Groom, Karol Hausman, Brian Ichter, Szymon
  Jakubczak, Tim Jones, Liyiming Ke, Sergey Levine, Adrian Li-Bell, Mohith
  Mothukuri, Suraj Nair, Karl Pertsch, Lucy~Xiaoyang Shi, Laura Smith, James
  Tanner, Quan Vuong, Anna Walling, Haohuan Wang, and Ury Zhilinsky.
\newblock {$\pi_0$: A Vision-Language-Action Flow Model for General Robot
  Control}.
\newblock In \emph{Proceedings of Robotics: Science and Systems},
  2025{\natexlab{b}}.

\bibitem[Chen et~al.(2026{\natexlab{a}})Chen, Zha, Liu, and Tevet]{foveavla}
Bingao Chen, Lihan Zha, C.~Karen Liu, and Guy Tevet.
\newblock {FoveaVLA}: Foveation actions replace the wrist camera in {VLAs}.
\newblock \emph{arXiv preprint arXiv:2610.20064}, 2026{\natexlab{a}}.

\bibitem[Chen et~al.(2026{\natexlab{b}})Chen, Yang, Tang, Huo, Lin, Wu, Liu,
  Chen, Zheng, Yuan, Li, Wang, Qi, Hu, Mei, Xuan, Yang, Zhu, Xu, Ma, and
  Chang]{abotm05}
Ronghan Chen, Yandan Yang, Zuojin Tang, Dongjie Huo, Tong Lin, Haoning Wu,
  Haoyun Liu, Yuzhi Chen, Lulu Zheng, Botai Yuan, Tianlun Li, Mingxin Wang,
  Dekang Qi, Bin Hu, Wei Mei, Yuze Xuan, Haolong Yang, Yanqing Zhu, Mu Xu,
  Zhiheng Ma, and Xinyuan Chang.
\newblock {ABot-M0.5: Unified Mobility-and-Manipulation World Action Model}.
\newblock arXiv preprint arXiv:2607.00678, 2026{\natexlab{b}}.

\bibitem[Chi et~al.(2023)Chi, Feng, Du, Xu, Cousineau, Burchfiel, and
  Song]{diffusionpolicy}
Cheng Chi, Siyuan Feng, Yilun Du, Zhenjia Xu, Eric Cousineau, Benjamin~CM
  Burchfiel, and Shuran Song.
\newblock {Diffusion Policy: Visuomotor Policy Learning via Action Diffusion}.
\newblock In \emph{Proceedings of Robotics: Science and Systems}, 2023.

\bibitem[Chuang et~al.(2026)Chuang, Zou, Lee, Gao, and Soltani]{lookfocusact}
Ian Chuang, Jinyu Zou, Andrew Lee, Dechen Gao, and Iman Soltani.
\newblock Look, focus, act: Efficient and robust robot learning via human gaze
  and foveated vision transformers.
\newblock In \emph{IEEE International Conference on Robotics and Automation
  (ICRA)}, 2026.

\bibitem[Cui et~al.(2026)Cui, Hu, Long, Li, Zhao, and Ren]{atvla}
Jin Cui, Yanbin Hu, Xinyue Long, Linkai Li, Boran Zhao, and Pengju Ren.
\newblock Look where it matters: Adaptive visual refinement for
  vision-language-action models.
\newblock arXiv preprint arXiv:2608.02197, 2026.

\bibitem[Dalal et~al.(2026)Dalal, Vashishtha, Mishra, Kim, Kanda, Ha, Lazebnik,
  Ji, and Jain]{attwarp}
Dwip Dalal, Gautam Vashishtha, Utkarsh Mishra, Jeonghwan Kim, Madhav Kanda,
  Hyeonjeong Ha, Svetlana Lazebnik, Heng Ji, and Unnat Jain.
\newblock Constructive distortion: Improving {MLLMs} with attention-guided
  image warping.
\newblock In \emph{International Conference on Learning Representations}, 2026.

\bibitem[Ehteshami~Bejnordi et~al.(2022)Ehteshami~Bejnordi, Habibian, Porikli,
  and Ghodrati]{salisa}
Babak Ehteshami~Bejnordi, Amirhossein Habibian, Fatih Porikli, and Amir
  Ghodrati.
\newblock {SALISA}: Saliency-based input sampling for efficient video object
  detection.
\newblock In \emph{European Conference on Computer Vision}, 2022.

\bibitem[Guo et~al.(2026)Guo, Liang, Wang, Li, Ma, Wu, Wang, Yu, Chen, Zhang,
  Xu, Zhang, Zhou, and Lu]{loopwam}
Wenxuan Guo, Yingping Liang, Zhaoguo Wang, Ziyuan Li, Angyuan Ma, Zhenyu Wu,
  Chengkun Wang, Bingyao Yu, Ze Chen, Borui Zhang, Xiuwei Xu, Meng Zhang, Erjin
  Zhou, and Jiwen Lu.
\newblock {LoopWAM: Closed-Loop Generalist Policy Improvement from Deployment
  Rollouts}.
\newblock arXiv preprint arXiv:2610.18873, 2026.

\bibitem[Kim et~al.(2026)Kim, Jang, Koo, Jang, Kim, Kim, Yoon, Jang, Choi, Han,
  Lee, Kwon, Jeon, Kang, Bae, Lee, Lee, Won, Ahn, Park, Sung, Lee, Han, Yoon,
  Joo, Son, Park, Cho, Moon, Kim, Dong, Cho, Kim, Kim, Kim, Kim, Lee, Ahn, Ryu,
  Choi, Shin, Jung, Kim, Kim, Chang, Kim, Park, Park, Cho, Park, Lee, Lee, Kim,
  Choe, Bhadu, Oh, Kim, Kim, Shim, Kim, Lee, Ka, Yang, Jung, Shukla, Lee, Bae,
  and Shin]{rldx1}
Dongyoung Kim, Huiwon Jang, Myungkyu Koo, Suhyeok Jang, Taeyoung Kim, Beomjun
  Kim, Byungjun Yoon, Changsung Jang, Daewon Choi, Dongsu Han, Donguk Lee,
  Heeseung Kwon, Hojin Jeon, Jaehyun Kang, Jaekyoung Bae, Jihyuk Lee, Jimin
  Lee, John Won, Joonwoo Ahn, Junhyeong Park, Junyoung Sung, Kyungmin Lee,
  Minseong Han, Minsung Yoon, Sejune Joo, Seonil Son, Seungcheol Park,
  Seunggeun Cho, Seungjun Moon, Seungku Kim, Yonghoon Dong, Yongjin Cho,
  Youngchan Kim, Chang~Hwan Kim, Dohyeon Kim, Heecheol Kim, Heewon Lee, Hensen
  Ahn, Hyungkyu Ryu, Hyunsoo Choi, Hyunsoo Shin, Jaeheon Jung, Jaewoo Kim,
  Jinwook Kim, Joochul Chang, Joonsoo Kim, Junghun Park, Jungwoo Park, Junho
  Cho, Junhyeok Park, Junwon Lee, Kangwook Lee, Kwanghoon Kim, Kyoungwhan Choe,
  Manoj Bhadu, Nayoung Oh, Sangjun Kim, Sangwoo Kim, Seunghoon Shim, Seunghyun
  Kim, Seungjun Lee, Seungyup Ka, Sungryol Yang, Wook Jung, Yashu Shukla,
  Yeonjae Lee, Yeonwoo Bae, and Jinwoo Shin.
\newblock {RLDX-1 Technical Report}.
\newblock arXiv preprint arXiv:2605.03269, 2026.

\bibitem[Kim et~al.(2025{\natexlab{a}})Kim, Finn, and Liang]{openvlaoft}
Moo~Jin Kim, Chelsea Finn, and Percy Liang.
\newblock Fine-tuning vision-language-action models: Optimizing speed and
  success.
\newblock In \emph{Proceedings of Robotics: Science and Systems},
  2025{\natexlab{a}}.

\bibitem[Kim et~al.(2025{\natexlab{b}})Kim, Pertsch, Karamcheti, Xiao,
  Balakrishna, Nair, Rafailov, Foster, Sanketi, Vuong, Kollar, Burchfiel,
  Tedrake, Sadigh, Levine, Liang, and Finn]{openvla}
Moo~Jin Kim, Karl Pertsch, Siddharth Karamcheti, Ted Xiao, Ashwin Balakrishna,
  Suraj Nair, Rafael Rafailov, Ethan~P. Foster, Pannag~R. Sanketi, Quan Vuong,
  Thomas Kollar, Benjamin Burchfiel, Russ Tedrake, Dorsa Sadigh, Sergey Levine,
  Percy Liang, and Chelsea Finn.
\newblock {OpenVLA}: An open-source vision-language-action model.
\newblock In \emph{Proceedings of the 8th Conference on Robot Learning}, pages
  2679--2713, 2025{\natexlab{b}}.

\bibitem[Li et~al.(2026)Li, Song, Zhao, Wang, Ding, Wang, Zeng, and
  Li]{spatialforcing}
Fuhao Li, Wenxuan Song, Han Zhao, Jingbo Wang, Pengxiang Ding, Donglin Wang,
  Long Zeng, and Haoang Li.
\newblock {Spatial Forcing: Implicit Spatial Representation Alignment for
  Vision-language-action Model}.
\newblock In \emph{International Conference on Learning Representations}, 2026.

\bibitem[Lin et~al.(2026)Lin, Ding, Wang, Zhuang, Liu, Tong, Song, Lyu, Huang,
  and Wang]{hifvla}
Minghui Lin, Pengxiang Ding, Shu Wang, Zifeng Zhuang, Yang Liu, Xinyang Tong,
  Wenxuan Song, Shangke Lyu, Siteng Huang, and Donglin Wang.
\newblock {HiF-VLA}: Hindsight, insight and foresight through motion
  representation for vision-language-action models.
\newblock In \emph{Proceedings of the IEEE/CVF Conference on Computer Vision
  and Pattern Recognition}, 2026.

\bibitem[Liu et~al.(2026)Liu, Gu, Wang, Xue, and Fu]{activevla}
Zhenyang Liu, Yongchong Gu, Yikai Wang, Xiangyang Xue, and Yanwei Fu.
\newblock {ActiveVLA}: Injecting active perception into vision-language-action
  models for precise {3D} robotic manipulation.
\newblock In \emph{Proceedings of the IEEE/CVF Conference on Computer Vision
  and Pattern Recognition}, pages 8141--8151, 2026.

\bibitem[Ma et~al.(2026)Ma, Ma, and Ma]{lygwm}
TianLiang Ma, Ting Ma, and ZhongYuan Ma.
\newblock {LY-GWM-RoboCasa-Human300: Prediction-Guided Action Selection with a
  Graph Dynamics Model}.
\newblock arXiv preprint arXiv:2610.23274, 2026.

\bibitem[{Meituan Robotics Team} et~al.(2026){Meituan Robotics Team}, Wang,
  Zhao, Jin, Xu, Chen, Wan, Yang, Wang, Zhou, Hu, Ru, Zhang, Huang, Zhang, Liu,
  Zhou, Liu, Li, Zhang, Zhang, Chen, Cao, Cai, Zeng, Yang, Yi, Tan, Xu, Jia,
  Tang, Xie, Zheng, Chng, Huang, Ye, Qin, and Wang]{meituan}
{Meituan Robotics Team}, Caicheng Wang, Chen Zhao, Chengpeng Jin, Duo Xu,
  Fengjiao Chen, Guanglu Wan, Heqing Yang, Jia Wang, Jiacong Zhou, Jie Hu,
  Jiyuan Ru, Lingfeng Zhang, Linjiang Huang, Manyuan Zhang, Ningchu Liu, Shijie
  Zhou, Si Liu, Shuo Li, Tianyi Zhang, Xi Zhang, Xin Chen, Xuezhi Cao, Xunliang
  Cai, Yanbin Zeng, Yang Yang, Yang Yi, Yanli Tan, Yifu Xu, Yimeng Jia, Yuchen
  Tang, Yuchen Xie, Yuxiang Zheng, Zerwei Chng, Zilong Huang, Zipeng Ye, Ziyuan
  Qin, and Ziqin Wang.
\newblock {Meituan-Robotics-0: A Vision-Language-Action Foundation Model for
  Desktop Manipulation}.
\newblock arXiv preprint arXiv:2611.08577, 2026.

\bibitem[Nasiriany et~al.(2026)Nasiriany, Nasiriany, Maddukuri, and
  Zhu]{robocasa365}
Soroush Nasiriany, Sepehr Nasiriany, Abhiram Maddukuri, and Yuke Zhu.
\newblock {RoboCasa365}: A large-scale simulation framework for training and
  benchmarking generalist robots.
\newblock In \emph{International Conference on Learning Representations}, 2026.

\bibitem[{NVIDIA}(2025{\natexlab{a}})]{groot15}
{NVIDIA}.
\newblock {NVIDIA Isaac GR00T N1.5: An Improved Open Foundation Model for
  Generalist Humanoid Robots}.
\newblock Official model release, 2025{\natexlab{a}}.
\newblock Version-specific model documentation; accessed November 9, 2026.

\bibitem[{NVIDIA}(2025{\natexlab{b}})]{groot16}
{NVIDIA}.
\newblock {NVIDIA Isaac GR00T N1.6-3B Model Card}.
\newblock Official model release, 2025{\natexlab{b}}.
\newblock Version-specific model documentation; accessed November 9, 2026.

\bibitem[Oquab et~al.(2024)Oquab, Darcet, Moutakanni, Vo, Szafraniec, Khalidov,
  Fernandez, Haziza, Massa, El-Nouby, Assran, Ballas, Galuba, Howes, Huang, Li,
  Misra, Rabbat, Sharma, Synnaeve, Xu, Jegou, Mairal, Labatut, Joulin, and
  Bojanowski]{dinov2}
Maxime Oquab, Timoth{\'e}e Darcet, Th{\'e}o Moutakanni, Huy~V. Vo, Marc
  Szafraniec, Vasil Khalidov, Pierre Fernandez, Daniel Haziza, Francisco Massa,
  Alaaeldin El-Nouby, Mahmoud Assran, Nicolas Ballas, Wojciech Galuba, Russell
  Howes, Po-Yao Huang, Shang-Wen Li, Ishan Misra, Michael Rabbat, Vasu Sharma,
  Gabriel Synnaeve, Hu Xu, Herv{\'e} Jegou, Julien Mairal, Patrick Labatut,
  Armand Joulin, and Piotr Bojanowski.
\newblock {DINOv2}: Learning robust visual features without supervision.
\newblock \emph{Transactions on Machine Learning Research}, 2024.

\bibitem[{Phasor}(2026)]{phasorm7}
{Phasor}.
\newblock {Phasor-m7: RoboCasa365 Benchmark Submission}.
\newblock Official RoboCasa365 leaderboard submission, 2026.
\newblock Submitted 2026-09-15; accessed November 9, 2026.

\bibitem[{Primotion}(2026)]{paimon0}
{Primotion}.
\newblock {Paimon-0: RoboCasa365 Benchmark Submission}.
\newblock Official RoboCasa365 leaderboard submission, 2026.
\newblock Submitted 2026-09-11; accessed November 9, 2026.

\bibitem[Recasens et~al.(2018)Recasens, Kellnhofer, Stent, Matusik, and
  Torralba]{learningtozoom}
Adria Recasens, Petr Kellnhofer, Simon Stent, Wojciech Matusik, and Antonio
  Torralba.
\newblock Learning to zoom: A saliency-based sampling layer for neural
  networks.
\newblock In \emph{European Conference on Computer Vision}, 2018.

\bibitem[Ren et~al.(2026)Ren, Yan, Zhang, He, and Li]{fisheyevla}
Ziang Ren, Zike Yan, Raymond Zhang, Xuguo He, and Zhongyu Li.
\newblock {Fisheye-VLA}: Decoupling coverage and acuity for manipulation with a
  single fisheye camera.
\newblock arXiv preprint arXiv:2609.25750, 2026.

\bibitem[{RoboCasa Team}(2026)]{robocasa_leaderboard}
{RoboCasa Team}.
\newblock {RoboCasa365} multi-task learning leaderboard.
\newblock \url{https://robocasa.ai/leaderboard}, 2026.
\newblock Updated November 1, 2026; accessed November 3, 2026.

\bibitem[Shi et~al.(2026)Shi, Xie, Liu, Sun, Liu, Wang, Zhou, Fan, Zhang, and
  Huang]{memoryvla}
Hao Shi, Bin Xie, Yingfei Liu, Lin Sun, Fengrong Liu, Tiancai Wang, Erjin Zhou,
  Haoqiang Fan, Xiangyu Zhang, and Gao Huang.
\newblock {MemoryVLA}: Perceptual-cognitive memory in vision-language-action
  models for robotic manipulation.
\newblock In \emph{International Conference on Learning Representations}, 2026.

\bibitem[Song et~al.(2026)Song, Zhou, Zhao, Chen, Ding, Yan, Huang, Tang, Wang,
  and Li]{reconvla}
Wenxuan Song, Ziyang Zhou, Han Zhao, Jiayi Chen, Pengxiang Ding, Haodong Yan,
  Yuxin Huang, Feilong Tang, Donglin Wang, and Haoang Li.
\newblock {ReconVLA}: Reconstructive vision-language-action model as effective
  robot perceiver.
\newblock In \emph{Proceedings of the AAAI Conference on Artificial
  Intelligence}, pages 18549--18557, 2026.

\bibitem[Thavamani et~al.(2021)Thavamani, Li, Cebron, and Ramanan]{fovea}
Chittesh Thavamani, Mengtian Li, Nicolas Cebron, and Deva Ramanan.
\newblock {FOVEA}: Foveated image magnification for autonomous navigation.
\newblock In \emph{Proceedings of the IEEE/CVF International Conference on
  Computer Vision}, pages 15539--15548, 2021.

\bibitem[Wang et~al.(2026)Wang, Xie, Jia, Xu, Zhang, and Yu]{patchwam}
Tianheng Wang, Zhou Xie, Heng Jia, Jianhua Xu, Tong Zhang, and Kaicheng Yu.
\newblock {An Action Is Worth One Patch: Unified World-Action Modeling with
  PatchWAM}.
\newblock arXiv preprint arXiv:2609.25961, 2026.

\bibitem[{World Agents}(2026)]{worlddreamer}
{World Agents}.
\newblock {WorldDreamer: RoboCasa365 Benchmark Submission}.
\newblock Official RoboCasa365 leaderboard submission, 2026.
\newblock Submitted 2026-06-20; accessed November 9, 2026.

\bibitem[Xiao et~al.(2026)Xiao, Li, Gao, Ye, Jin, Qian, Zhang, Wu, and
  Yu]{avavla}
Lei Xiao, Jifeng Li, Juntao Gao, Feiyang Ye, Yan Jin, Jingjing Qian, Jing
  Zhang, Yong Wu, and Xiaoyuan Yu.
\newblock {AVA-VLA}: Improving vision-language-action models with active visual
  attention.
\newblock In \emph{Proceedings of the IEEE/CVF Conference on Computer Vision
  and Pattern Recognition}, pages 13453--13463, 2026.

\bibitem[{Xiaomi Robotics Team} et~al.(2026){Xiaomi Robotics Team}, Guo, Jin,
  Li, Li, Li, Liu, Peng, Qin, Su, Sun, Sun, Suo, Wang, Wang, Wu, Xia, Zhang,
  Zhao, Chen, Chen, He, Li, Li, Li, Qu, Song, Xiang, Xie, Xu, Ye, Ye, Zhao, and
  Zhou]{xiaomi}
{Xiaomi Robotics Team}, Jun Guo, Piaopiao Jin, Jason Li, Peiyan Li, Yingyan Li,
  Futeng Liu, Wanli Peng, Optimus Qin, Yifei Su, Nan Sun, Qiao Sun, Runze Suo,
  Heyun Wang, Yunhong Wang, Rujie Wu, Caoyu Xia, Lina Zhang, Jack Zhao,
  Guoliang Chen, Wenlong Chen, Xinze He, Bin Li, Qing Li, Zhuorong Li, Heng Qu,
  Wenxuan Song, Diyun Xiang, Yifan Xie, Peiran Xu, Hangjun Ye, Wen Ye, Han
  Zhao, and Quanyun Zhou.
\newblock {Xiaomi-Robotics-1: Scaling Vision-Language-Action Models with over
  100K Hours of Real-World Trajectories}.
\newblock arXiv preprint arXiv:2607.15330, 2026.

\bibitem[Xu et~al.(2025)Xu, Wang, Xia, Zhu, Huang, and Xu]{vlacache}
Siyu Xu, Yunke Wang, Chenghao Xia, Dihao Zhu, Tao Huang, and Chang Xu.
\newblock {VLA-Cache}: Efficient vision-language-action manipulation via
  adaptive token caching.
\newblock In \emph{Advances in Neural Information Processing Systems}, 2025.

\bibitem[Ye et~al.(2026)Ye, Wang, Ni, Huang, Zhao, Li, Li, Li, Lv, Liu, Cao,
  Li, Deng, Mei, Wang, Chen, Zhou, Wang, Chang, Li, Zhou, Ye, Liu, and
  Zhu]{gigaworld}
Angen Ye, Boyuan Wang, Chaojun Ni, Guan Huang, Guosheng Zhao, Hao Li, Hengtao
  Li, Jie Li, Jindi Lv, Jingyu Liu, Min Cao, Peng Li, Qiuping Deng, Wenjun Mei,
  Xiaofeng Wang, Xinze Chen, Xinyu Zhou, Yang Wang, Yifan Chang, Yifan Li,
  Yukun Zhou, Yun Ye, Zhichao Liu, and Zheng Zhu.
\newblock {GigaWorld-Policy: An Efficient Action-Centered World--Action Model}.
\newblock arXiv preprint arXiv:2603.17240, 2026.

\bibitem[Zhai et~al.(2023)Zhai, Mustafa, Kolesnikov, and Beyer]{siglip}
Xiaohua Zhai, Basil Mustafa, Alexander Kolesnikov, and Lucas Beyer.
\newblock Sigmoid loss for language image pre-training.
\newblock In \emph{Proceedings of the IEEE/CVF International Conference on
  Computer Vision}, 2023.

\bibitem[Zhang et~al.(2026{\natexlab{a}})Zhang, An, Frost, Wang, Liu, Zhang,
  Drozdzal, and Bar]{prowam}
Fei Zhang, Zhaochong An, Duncan Frost, Yikai Wang, Pengfei Liu, Ya Zhang,
  Michal Drozdzal, and Amir Bar.
\newblock {World Action Modeling with Progressive Visual Planning}.
\newblock arXiv preprint arXiv:2610.02508, 2026{\natexlab{a}}.

\bibitem[Zhang et~al.(2026{\natexlab{b}})Zhang, Zhao, Fan, Yan, Li, Tang, Fu,
  Wu, Weng, Zhang, Li, Zhang, Bai, and Li]{prts}
Yang Zhang, Jiangyuan Zhao, Chenyou Fan, Fangzheng Yan, Tian Li, Haitong Tang,
  Sen Fu, Xuan'er Wu, Qizhen Weng, Weinan Zhang, Xiu Li, Chi Zhang, Chenjia
  Bai, and Xuelong Li.
\newblock {PRTS: A Primitive Reasoning and Tasking System via Contrastive
  Representations}.
\newblock In \emph{Conference on Robot Learning}, 2026{\natexlab{b}}.

\bibitem[Zheng et~al.(2025)Zheng, Liang, Huang, Gao, Daum{\'e}, Kolobov, Huang,
  and Yang]{tracevla}
Ruijie Zheng, Yongyuan Liang, Shuaiyi Huang, Jianfeng Gao, Hal Daum{\'e}, III,
  Andrey Kolobov, Furong Huang, and Jianwei Yang.
\newblock {TraceVLA}: Visual trace prompting enhances spatial-temporal
  awareness for generalist robotic policies.
\newblock In \emph{International Conference on Learning Representations}, 2025.

\end{thebibliography}
